\documentclass[runningheads]{llncs}
\usepackage{graphicx}
\usepackage{comment}
\usepackage{amsmath,amssymb} % define this before the line numbering.
\usepackage{color}
\usepackage{soul}
\usepackage{array}
\usepackage{booktabs}
\usepackage{hyperref}
\hypersetup{
    colorlinks=true,
    linkcolor=blue,
    filecolor=magenta,      
    urlcolor=cyan,
}
\graphicspath{{Images/}}
\usepackage{soul,color}
\usepackage{subcaption}
\begin{document}
% \linenumbers
\pagestyle{headings}
\mainmatter
\def\ECCVSubNumber{1395}  % Insert your submission number here

\title{DPDist : Comparing Point Clouds Using Deep Point Cloud Distance} % Replace with your title

\begin{comment}
\titlerunning{ECCV-20 submission ID \ECCVSubNumber} 
\authorrunning{ECCV-20 submission ID \ECCVSubNumber} 
\author{Anonymous ECCV submission}
\institute{Paper ID \ECCVSubNumber}
\end{comment}
%******************

% \begin{comment}
% \titlerunning{DPDist : Comparing Point Clouds Using Deep Point Cloud Distance}
\author{Dahlia Urbach\inst{1} \and
Yizhak Ben-Shabat\inst{2} \and
Michael Lindenbaum\inst{1}}
\authorrunning{D. Urbach, Y. Ben-Shabat, M. Lindenbaum}
% \authorrunning{D. Urbach, et al.}

\institute{Techion IIT, Israel \and 
Australian National University, Australian Centre for Robotic Vision, Australia \\
\email{dahliau@technion.ac.il, yizhak.benshabat@anu.edu.au, mic@cs.technion.ac.il}\\
\url{https://github.com/dahliau/DPDist}}
% \end{comment}
%******************
\maketitle

\begin{abstract}
We introduce a new deep learning method for point cloud comparison.  Our approach, named Deep Point Cloud Distance (DPDist),   measures the distance between the points in one cloud and the estimated surface from which the other point cloud is sampled. 
The surface is estimated locally using the 3D modified Fisher vector representation. The local representation reduces the complexity of the surface, enabling effective learning, which generalizes well between object categories. 
We test the proposed distance in challenging tasks, such as similar object comparison and registration, and show that it provides significant improvements over commonly used distances such as Chamfer distance,  Earth mover's distance, and others. 
% Our approach introduces a conceptual change that enables measuring the distances and compute the gradients between two surfaces represented by 3D point clouds directly from their raw points (and backpropagate).
\keywords{ 3D Point Clouds, 3D Computer Vision, 3D Deep Learning,  Distance, Registration
% We would like to encourage you to list your keywords within the abstract section
}
\end{abstract}

\section{Introduction}
 
Recent advancements in 3D sensor technology have led to the integration of 3D sensors into many application domains, such as virtual and augmented reality, robotic vision, and autonomous systems. These sensors supply a set of 3D points, sampled on surfaces in the scene, known as a point cloud.
As raw, memory-efficient outputs of 3D sensors, point clouds are a common 3D data representation. 
However, in contrast to more traditional data (e.g., images), point clouds are unstructured, unordered, and may have a varying number of points. Therefore, unlike traditional signals, they may not be represented as values on some regular grid, and are difficult to process using common signal processing tools.

In particular, many applications, such as registration, retrieval, autoencoding etc., require comparisons between two or more point clouds. Comparing point clouds is difficult for two main reasons. First, because they are not a function on a grid, point clouds cannot be compared using a common metric (such as Euclidean metric). Second, when comparing point clouds sampled from a 3D surface, we usually want to compare their underlying surfaces and not the given sample points. 
Common methods for point cloud comparisons, such as  Chamfer distance and Earth mover's distance (see more in Section \ref{Sec:related}) compare the given clouds directly, rely on the unstable  correspondence process, and are sensitive to sampling. 

\begin{figure*}[tp]
    \centering
    \includegraphics[width=0.98\linewidth]{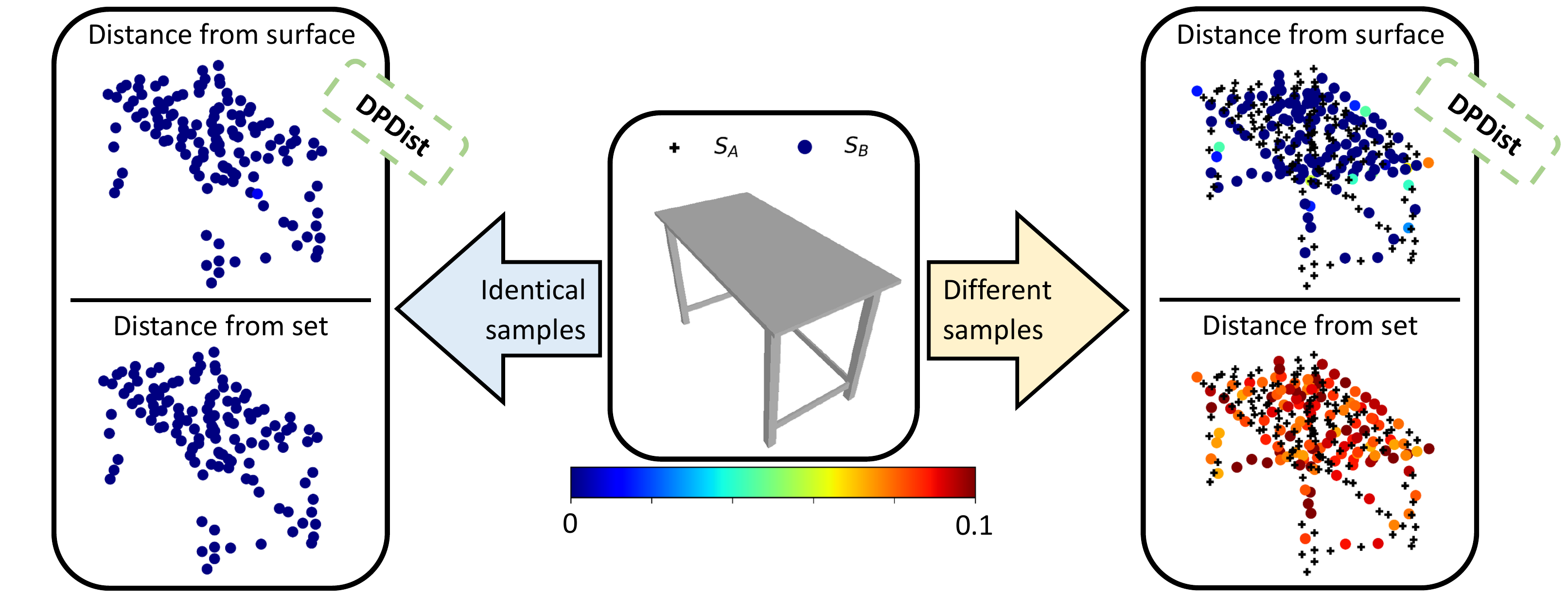}
    \caption{Distance to surface vs. distance to sampled point cloud.  Every figure contains two samples of the same object (a table). The samples can be identical (left) or different (right). The distance of every point from one point cloud (disks) to the surface estimated from the other point cloud ($+$) is given as a colored code in the top two figure. One can see that even for the different sampling most of the distances are  still small. The distance from every point in one point cloud (disks) to the nearest point in the other point cloud (+) is presented as a  color code in the bottom figure. We can see that different sampling produce significantly larger distances.}
    \label{fig:Teaser}
\end{figure*}

We propose a method for comparing point clouds that measures the distance between the surfaces that they were sampled on (see Fig. \ref{fig:Teaser}).
We use a deep learning network to estimate the distance function from a point to an underlying continuous surface. By using a vector representation of point clouds, the 3D modified Fisher vector (3DmFV) \cite{ben20183dmfv}, the  network can work  directly with point clouds, despite their irregular structure. 
An additional advantage of the particular 3DmFV representation is its grid structure, which enables the network to process a local surface representation rather than calculating a representation at every location. Working on local representation reduces the complexity of the surface, enabling efficient and effective learning, which generalizes well between object categories. 

We test the robustness of the proposed approach to sampling, on challenging tasks such as detecting small transformation and comparing similar objects. We incorporate it into the learning process of a registration network. We compare it to several, more traditional, distance measures and show its advantage, which is significant, especially when the task is harder and the point cloud is sparser. 

The main contributions of this paper are as follow: 
\begin{itemize}
    \item DPDist: A new distance measure between point clouds, that operates directly on raw unstructured points, but measures distance to the underlying continuous surface.
    \item An algorithm that implements the DPDist using local implicit representation, in an effective and efficient way. 
    \item An improved variant of the PCRNet registration algorithm \cite{sarode2019pcrnet} that applies the proposed DPDist in a training loss function.
    % We introduce a novel loss function for learning tasks. 
    % that uses the DPDist metric and show that it achieved state of the art results compared to Euclidean based metrics (EMD,CD) for the tasks of retrievals and registration.
\end{itemize}

% Our Tensorflow  implementation as well as the pretrained models are available at \url{https://github.com/dahliau/DPDist}.

%Teaser Image:

\section{Related Work}
\label{Sec:related}

\subsection{Point Cloud Distance }

Point clouds may represent raw data, but in the context of 3D sensors, they represent the surfaces on which they are sampled. As such, the distance between the point clouds, should preferably refers to the distance between the sampled continuous surfaces and be robust to sampling and noise.

Most distances between points sets are based on an underlying metric, assigning a distance value $\|a-b\|$ to every two points $a,b$. Like the majority of previous work, we use a Euclidean metric. 

Consider two point clouds $S_A, S_B \subseteq\mathbb{R}^3$ , with $N_A, N_B$ points in each cloud, respectively. 
An early method for comparing point sets is the Hausdorff distance (${\cal D}_H()$), which builds on the minimum distance from a point to a set 
\begin{equation}
  d(x,y) = \|x-y\|_2\\
\end{equation}
\begin{equation}
  D(x,S) = \min_{y \in S}d(x,y)
\end{equation}
% \begin{equation}
%   d(x,S) = \min_{y \in S}\|x-y\|_2\\
% \end{equation}
and calculates a symmetric max min distance
% \cite{some-nice-ref}
\cite{huttenlocher1993comparing}.  
% \cite{rucklidge1996efficient}.
% \ref{hausdorff}
\begin{equation}\label{hausdorff}
   {\cal D}_{H}(S_A,S_B) = \max\{\max_{a \in S_A}D(a, S_B), \max_{b \in S_B}D(b,S_A) \}
%   D_{H}(S_A,S_B) = \max\{\max_{a \in S_A}d(a, S_B), \max_{b \in S_B}d(b,S_A) \}
%     D_{H}(S_A,S_B) = \max\{\max_{a \in S_A}\min_{b \in S_B} (\|a-b\|_2) , \max_{b \in S_B}\min_{a \in S_A} (\|b-a\|_2)\}
\end{equation}
The Chamfer distance (${\cal D}_{CD}()$) (Eq. \ref{chamfer}), and partial Hausdorff (${\cal D}_{PH(f)}()$) (Eq. \ref{partial}) distances are two of its variants, based on averaging (instead of taking the maximum) and on robustly ignoring a fraction ($1-f$) of the points that are far from the other object, respectively
% \begin{equation}\label{chamfer}
%     D_{C}(S_A,S_B) = \frac{1}{N_A}\sum_{a\in S_A}d(a,S_B) +
%     \frac{1}{N_B}\sum_{b \in S_B} d(b, S_A) 
% \end{equation}
\begin{equation}\label{chamfer}
    {\cal D}_{CD}(S_A,S_B) = \frac{1}{N_A}\sum_{a\in S_A}\min_{y \in S_B}d(a,y)^2 +
    \frac{1}{N_B}\sum_{b \in S_B} \min_{y \in S_A}d(b,y)^2
    % D_{C}(S_A,S_B) = \frac{1}{N_A}\sum_{a\in S_A}\min_{y \in S_B}\|a-y\|_2^2 +
    % \frac{1}{N_B}\sum_{b \in S_B} \min_{y \in S_A}\|b-y\|_2^2
\end{equation}

% \begin{equation}\label{partial}
%    D_{PH(F)}(S_A,S_B) = \max\{T | \frac{|\{a |a \in S_A, \min_{b \in S_B} \|a,b\| \leq T \}|}{|S_A|}<F,   \frac{|\{b |b \in S_B, \min_{a \in S_A} \|a,b\| \leq T \}|}{|S_B|}<F \}
% \end{equation}

\begin{equation}\label{partial}
    {\cal D}_{PH(f)}(S_A,S_B) = \max\{T | \frac{|\{a | D(a,S_B) \leq T \}|}{|S_A|}<f,
     \frac{|\{b | D(b,S_A) \leq T \}|}{|S_B|}<f \}
    %   D_{PH(f)}(S_A,S_B) = \max\{T | \frac{|\{a | d(a,S_B) \leq T \}|}{|S_A|}<f,
    %  \frac{|\{b | d(b,S_A) \leq T \}|}{|S_B|}<f \}
\end{equation}
where $a \in S_A, b \in S_B$.
% ,and  $D(x,S) = \min_{y \in S}\|x-y\|_2$. 

Unlike the Hausdorff distance and its variant, which rely on finding the nearest neighbor to every point, the Earth mover's distance (${\cal D}_{EMD}$) (Eq. \ref{emd}), also known as the Wasserstein distance, is based on finding the 1-1 correspondence (or bijection, $\xi$) between the two point sets, so that the sum of distances between corresponding points is minimal: 
\begin{equation}\label{emd}
    {\cal D}_{EMD}(S_A,S_B) = \min_{\xi:S_A\xrightarrow{}S_B} \sum_{a\in S_A} \|a - \xi(a)\|_2
        % D_{EM}(S_A,S_B) = \min_{\phi:S_A\xrightarrow{}S_B} \sum_{a\in S_A} \|a - \Phi(a)\|_2
\end{equation}
Clearly, this distance measure is limited to point clouds with the same point count. 
Other measures includes Geodesic distances, which provide deformation insensitive measures  \cite{bronstein2011shape}, and measures relying on the distance between point cloud vector descriptors, such as PointNet \cite{qi2017pointnet},
VoxNet \cite{maturana2015voxnet}, or 3DmFV \cite{ben20183dmfv}. 

Here we focus on the direct distances, and specifically on CD and EMD like measures. These are the leading approaches for assessing subtle differences in point clouds, and are used as evaluation distances and as loss functions for the training of neural networks \cite{sarode2019pcrnet,fan2017point,achlioptas2017learning,yang2018foldingnet,groueix2018atlasnet,Li_2018_CVPR,Zhao_2019_CVPR}.
They highly depend on point correspondence, which makes them sensitive to sampling and noise.
In this work, we propose a novel method for comparing 3D point clouds where the measured distance is between the points from the first point cloud to the surface represented by the second (without constructing a mesh) and the other way round. 
As a distance, it can replace the aforementioned distances as a loss function for various neural networks that require point cloud comparison.

\subsection{Deep Learning on 3D Point Clouds}
Processing a point cloud with deep learning is challenging because point cloud representation  is unordered, unstructured, and has unknown number of points, rendering it an unnatural input for deep networks.  
Several methods were proposed to overcome these challenges.
%voxels:
One approach quantizes the points into a voxels grid. This approach allows to directly use 3D convolutional neural networks (CNNs) \cite{maturana2015voxnet} but induces quantization.
%KD
Another approach encodes the point cloud with a kd-tree and uses it to learn shared weights for nodes in the tree  \cite{klokov2017escape}.
%PN
A recent popular network for processing point clouds is PointNet \cite{qi2017pointnet}. It computes features separately for each input point, and then extracts a global feature using a permutation-independent (symmetric) function (e.g., max/avg). 
%PN++
An effective variant, PointNet++ \cite{qi2017pointnet++}, applies the PointNet encoding locally and hierarchically. 

%3DmFV
The recent 3D modified Fisher Vector (3DmFV) approach \cite{ben20183dmfv} builds on the well known Fisher Vectors \cite{sanchez2013image} which, in turn, are based on the Fisher Kernel (FK) principles \cite{jaakkola1999exploiting}. 3DmFV represents each point's deviations from a mixture of Gaussians, lying on a regular grid, and then applies symmetric functions to get a global generalized Fisher Vector representation \cite{sanchez2013image} that integrates into a 3D CNN.
% \cite{jaakkola1999exploiting}
% 3DmFV builds on the well known Fisher Vectors [14] which, in turn, are based on the Fisher Kernel (FK) principles* . It characterizes data samples of varying sizes by their deviation from a generative model, which, here, is a GMM.  It was shown that the FK are optimal in the sense that decisions based on it are  asymptotically as good as those made by the maximum a posteriori (MAP) decision rule for this model.
%
The proposed method uses the regular grid structure of the 3DmFV representation, to  extract  multiple local patches without any recalculation.
%IBS: Maybe this section is missing foldingnet, ppfnet, ppf-foldingnet, but maybe not since they are general extensions to pointnet. think about it. 

\subsection{Deep Implicit Function}
% \hl{Maybe add implicit surface approximation methods like least squares, poisson etc}.
Two recent works that are the closest to the proposed method rely on 
continuous implicit surface representations.
Occupancy Networks \cite{mescheder2019occupancy}
learn an indicator function between the inner part and the outer part of a model, where the boundary between these parts is an implicit function representing the continuous boundary. 
DeepSDF \cite{park2019deepsdf} learns a signed distance function to  the surface, and estimates the boundary as the zero surface. 
Both methods rely on a global presentation of the point cloud and the estimated surface. 
Occupancy Networks \cite{mescheder2019occupancy} rely on PointNet, and DeepSDF specifies each model's surface by an decoder training process, which provides both the weights and the latent layer values (without using an encoder).

Unlike DeepSDF,
% \cite{park2019deepsdf}, 
our proposed algorithm is able to transform a point cloud into underlying surface representation quickly and can therefore use the distance-to-surface principle to calculate the distance between point clouds. Unlike Occupancy Networks, our proposed method calculates the distance to the surface and can therefore estimate whether two point clouds, sampled from surfaces, are indeed sampled from the same surface. In addition, unlike the two aforementioned methods, our approach can extract and use effective local representation, which makes both the learning and the estimate, accurate and effective.

\subsection{Registration}
\label{rw:reg}

%IBS: Need to state and cite keypoint based registration approaches (PPF-Foldenet has this covered), while they are not 1005 relevant for this paper, if we are discussing registration in general they should appear as well. 
Registration between point clouds is a fundamental task. A popular, classic (non-learnable) algorithm is Iterative Closest Point (ICP) \cite{rusinkiewicz2001efficient}, which relies on finding closest point pairs. 
To overcome ICP's convergence to local minima, recent ICP variants
\cite{yang2015go} start by finding 
a good initialization at some computational expense. 

Recently, deep-learning-based registration methods have been introduced.
Some methods  
\cite{wang2019deep,aoki2019pointnetlk}
learn to regress the transformation parameters. The
PCRNet \cite{sarode2019pcrnet}, however, 
learns by comparing point clouds. The method aims to find the transformation that converts the measured point cloud (source) to the model (template). At every iteration, it maintains a temporary transformed source, calculated with a currently available transformation. A pose estimation network gets PointNet representations of this temporary cloud and of the template cloud, and provides a new, improved, transformation (represented as translation parameters and quaternion rotation parameters). The network is trained by a loss based on the Chamfer distance or the EMD between the temporary cloud and the template cloud.
In this paper, we test the proposed new distance DPDist, as a loss function for training the PCRNet.

\section{The Deep Distance}
We now present the main contribution of this work: a new method for fine comparison of point clouds that we call the Deep Point Cloud Distance (DPDist). 

The DPDist method is based on estimating the  distances of points from one cloud to the underlying continuous surface corresponding to the other point cloud. 

Let $A,B \in \mathbb{R}^3$ be two continuous surfaces, and $S_A=\{a_i\}_{i=1}^{N_A}, S_B=\{b_j\}_{j=1}^{N_B}$ 
two sets of points sampled from them.  
Our aim is to find the distance from each point $b_j \in S_B$ to the closest point in $A$ (and vice versa). 
Formally, this distance is
\begin{equation}
 D(b,A)=\min_{y \in A} d(b,y),
\end{equation}
where $D()$ is the distance from a point to a surface and $d()$ is any distance between points in $\mathbb{R}^3$ (here we shall use the Euclidean distance).  The surface $A$ is not available, and hence we propose to use the approximation
\begin{equation}
 \hat{D}(b,A)=\phi(b,S_A) ,
\end{equation}
 where $\phi$ is a learned distance function depending on the samples of $A$. 
 
In principle, we can design a network that accepts the coordinates of the point $b$, and some representation of the cloud of points $S_A$ (e.g., PointNet \cite{qi2017pointnet} or 3DmFV \cite{ben20183dmfv}). This network should learn to provide the distance.  
We found, however that due to the large variation of objects, learning the distance function is hard. In particular, even with a complex network and a lot of examples, the results we obtained were not very good. See \cite{park2019deepsdf,mescheder2019occupancy} where a similar task of estimating an implicit function by using at least 7 fully connected layers to express the complicated geometric features.
% They use at least seven fully connected layers because they try to infer complicated geometric details from a global representation.

\subsection{Local Representation}
\begin{figure*}[t]
    \centering
    \includegraphics[width=12cm]{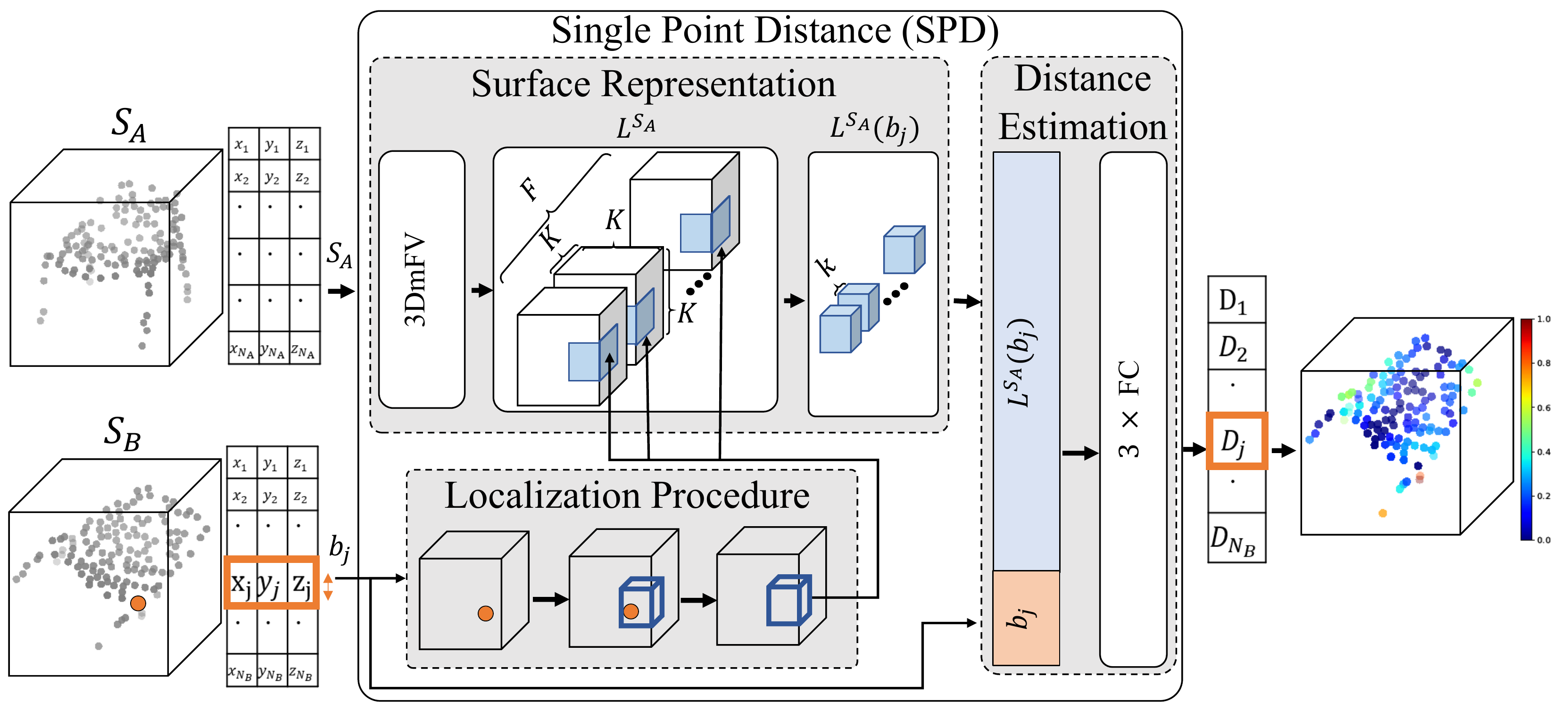}
    \caption{SPD illustration. Two point clouds ($S_A,S_B$) are input to a network that estimates the distance between each query point from $S_B$ to the surface represented by $S_A$. SPD uses 3DmFV to extract surface representation $L^{S_A}$ (4D tensor) from $S_A$. It then applies a localization procedure for each query point $b_j$ to find its closest grid point (the grid is distributed uniformly: $K^3$). Given the grid point, it extracts the corresponding 4D sub tensor $L^{S_A}(b_j)$. 
    The SPD processes each query point $b_j$ and its matched local representation to produce its distance from the surface $D_j=D_{SPD}(b_j,S_A)$.}
    \label{fig:SPD}
\end{figure*}
Because of the difficulty of learning the distance function, we prefer to model the surface $A$ by parts, so that every part is less geometrically complex. In principle, we could derive such a piecewise model by training a different distance estimator for every spatial region. Instead, we choose the more elegant approach of using the 3D modified Fisher Vector (3DmFV) representation.

% We choose to use the 3D modified Fisher Vector (3DmFV) representation. 
The 3DmFV representation is preferable for two reasons. First, it provides better performance than some other representations (and in particular, it is better than the PointNet; see \cite{ben20183dmfv}). In addition, it uses a grid structure therefore allowing us to specify a local, partial representation by choosing a subgrid. 

First, we calculate the global representation from the cloud $S_A$. To that end, we decide on the representation grid size $K$, and specify the 3D grid of size $K\times{K}\times{K}$. This grid specifies a mixture of Gaussians, one for each grid point. The representation itself results from calculating the derivative of the Gaussian with respect to its parameters, and taking maximum, minimum, and average statistics over these derivatives; see \cite{sanchez2013image,ben20183dmfv}. Overall, we calculate $F$ statistics for each Gaussian and concatenate them to a global representation $L^{S_A} \in 
\mathbb{R}^{K\times{K}\times{K}\times{F}}$. 
This representation is calculated once for every point cloud and the partial representations are simply cut from it. 

We extract the local representation from this global representation as follows:  for each query point $b_j$ from the point cloud $S_B$,  find its nearest grid point and specify a subgrid of size $k$, centered at this grid point ($k$ is always odd). Then, the entries of $L^{S_A}$ corresponding to the subgrid are extracted and concatenated to give the local representation, denoted  $L^{S_A}(b_j) \in 
\mathbb{R}^{k\times{k}\times{k}\times{F}}
$. Note that the local representation is a representation of the point cloud $S_A$, in a region determined by $b_j$.

\begin{figure*}[t]
    \centering
    \includegraphics[width=1\textwidth]{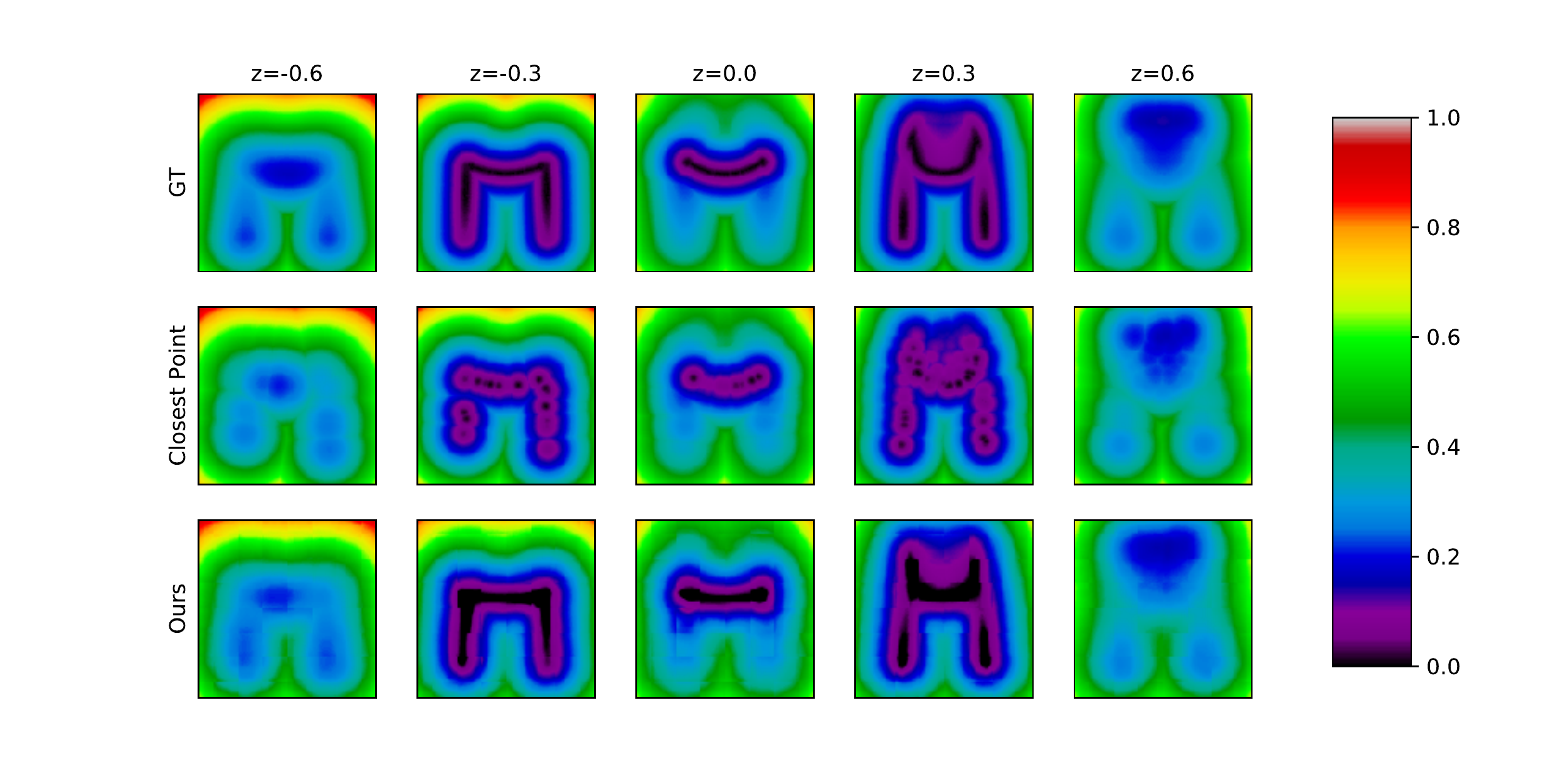}
    \caption{Three distance map to the "Chair" CAD model:  The ground truth distance from the CAD model (upper row), the distance to the 'closest point' in a 128 point cloud, sampled from the CAD model (middle row), and the proposed SPD network distance from the surface specified by the same point cloud (bottom row). For each 3D distance map, several XY slices 
    % corresponding to   $z=\{-0.6,-0.3,0,0.3,0.6\}$
    are shown.  
    Clearly, the SPD is smoother than the distance to the point cloud, and approximates the ground truth better.
    }
    \label{fig:Distance_maps128}
\end{figure*}
\subsection{Point-to-Implicit-Surface Distance Estimation}
To estimate the distance from a point to the underlying surface associated with a point cloud, we use a learned function. We chose a multilayer, fully connected neural network (FC). The input to this network is the coordinate vector specifying the point $b_j$, concatenated to the local representation $L^{S_A}(b_j)$. The neural network, denoted $\psi$, learns to estimate the distance  $\psi(b_j,L^{S_A}(b_j))$ 
between the  point and the corresponding local representation.
% The full process, which gets the point $b_j$, extract the local representation corresponding to $b_j$ from the 3DmFV vector, concatenates them, and processes them using the network $\psi$, is denoted SPD (Single point distance) network.  
%
The full process, which gets the point $b_j$ and the point cloud $S_A$ (as input), is carried out as follows. 
The process first extracts the local representation corresponding to $b_j$ from the 3DmFV vector; concatenates it to the point coordinates; and processes them using the network $\psi$.
We denote it: SPD (Single point distance).
That is
\begin{equation}
 \hat{D}(b_j,A)=  D_{SPD}(b_j,S_A)=
  \psi(b_j,L^{S_A}(b_j))
%  \hat{D}(b_j,A)=  {SPD}(b_j,L^{S_A}(b_j))
\end{equation}

The detailed description of the SPD neural network is presented in Fig. \ref{fig:SPD}. We found indeed that learning the distance function of a local representation is faster, more accurate, and can generalize better for unseen objects. Furthermore, representing a surface $A$  directly from a set of points $S_A$ using the proposed approach yields sampling invariance (sparse or non-uniform point clouds may represent the same surface). 
% We illustrate the SPD distance maps in Fig. \ref{fig:Distance_maps128}, we can see that SPD performs completion for low-resolution point clouds (e.g $N=128$). 
Fig. \ref{fig:Distance_maps128}, illustrates the distance maps of SPD and compares it to the distance to the cloud, for a dense set of query points that lie on a regular XY grid in different spatial depths ($z=\{-0.6,-0.3,0,0.3,0.6\})$. The advantage of the SPD for this low sampling data ($N=128$) is clear. 

\subsection{Estimating the Distance Between Point Clouds}
The point to surface distance estimates, developed above may be easily applied to measuring the distance between two point clouds. The average distance between all points in one cloud, $S_B$, to the underlying surface corresponding to the other point cloud is a straightforward choice. Here we often use a symmetrized version to yield the DPDist distance between two point clouds.

\begin{equation}\label{DPDist_eq}
    {\cal D}_{DPDist}(A,B) = \frac{1}{N_A}\sum_{i=1}^{N_A}{D_{SPD}(a_i,S_B)}+
    \frac{1}{N_B}\sum_{j=1}^{N_B}{D_{SPD}(b_j,S_A)}
    % DPDist(A,B) = \frac{1}{N_A}\sum_{i=1}^{N_A}{SPD(a_i,L^{S_B}(a_i))}+
    % \frac{1}{N_B}\sum_{j=1}^{N_B}{SPD(b_j,L^{S_A}(b_j))}
\end{equation}

\subsection{DPDist as Loss Function for Training Neural Networks}
After DPDist is fully trained, it can be used as a distance estimator building block that can be connected as a loss function to various tasks. Generally, its weights can either be frozen or adjusted to the desired task, but as a loss function, the weights remain constant. 
It takes two point clouds $S_A$,$S_B$ as input and outputs the estimated distance between the underlying represented surfaces.

DPDist uses a neural network, and is therefore differentiable and can be easily integrated as a loss function for training in different tasks.

\section{Experiments}
We start by  conducting a thorough analysis of the method's robustness to sampling. We would like to emphasize that for tasks involving coarse differences between point clouds, such as categorization and coarse registration, distances that are based directly on the point clouds (e.g., Chamfer distance) or even distances based on vector representations (e.g., distance between 3DmFVs) should suffice. The distance suggested here can detect small, subtle differences. Therefore, we chose to experiment with the following challenging tasks: object instance identification, detecting small translations, and detecting small rotations.
We compare our method with the following  methods: Hausdorff (H), Chamfer distance (CD), partial Hausdorff (PHx), Earth mover's distance (EMD), and 3DmFV representation (we have shortened PH(f) to PHx, where f$=0.$x). 
We then evaluate our method's effectiveness as a loss function in training a registration network. 
We provide additional point cloud generation experiments in the supplemental material. They show that DPDist's main advantage of sampling invariance can introduces significant challenges for this task. Additionally, we report in the supplemental results on real-world data that align with our findings on CAD models. 

\subsection{Setup Details}
\subsubsection{Dataset}
The experiments were conducted on the ModelNet40 dataset \cite{Wu_2015_CVPR}. It contains 12311 CAD models from 40 object categories.
We normalize each shape to $80\%$ of the unit sphere.
We split the data into train/test as in  \cite{qi2017pointnet}.

\subsubsection{Training data}
For training the SPD network, we sample points from the CAD models, keeping their distance from the surface. At each training step, we sample two point clouds. The first, representing the surface, contains $N$ points sampled from the surface. The second set is the query set. It contains $0.5N$ points sampled from the surface, $0.25N$ points sampled uniformly in the region closer than $0.1$ to the surface, and $0.25N$ points uniformly sampled from the unit cube. 
Adding the second type of query points increases the density of the query points near the surface and enhances the learning of small geometric details; see \cite{park2019deepsdf}). 

\subsubsection{Training details}
Fig. \ref{fig:SPD} describes the overall architecture of SPD:
We set the 3DmFV layer parameters to $K=8$, $k=5$, and the Gaussian's sigma to $0.125$. Additional ablation study of the influence of 3DmFV parameters is provided in the supplemental material. 
To evaluate the parameters we used marching cubes  \cite{lorensen1987marching} for surface reconstruction as done in \cite{park2019deepsdf,mescheder2019occupancy}.
% material
The neural network is composed of three fully connected layers with a size of 1024, each processed with RELU activation.
During training, we minimized the mean $L_1$ distance between the network output (our prediction) and the real distance (GT) associated with each query point,  i.e. 
$Loss = \frac{1}{|S_B|}\sum_{x_i \in S_B}||D_{SPD}(x_i, S_A)-GT(x_i)||_1$ .

We train separate networks for each point cloud input size  $N=32, 64, 128, \allowbreak 256, 512, 1024$.
Our early experiments showed that we could use a network trained for an input size of $N=512$ for all different size inputs; however, training each network for a specific input size yields higher accuracy.

For the following experiments, we trained the SPD using the train-set of the "Chair" category, which contains 889 CAD models.For specific training parameters and further details, please refer to the supplemental material.

\subsubsection{Evaluation data sampling}
To obtain the data for evaluation, we first sample $N_C$ points uniformly from each CAD model using Farthest Point Sampling (FPS) \cite{eldar1997farthest}. We then sample two disjoint sets $S_A, S_B$ with $N_A,N_B$. In most experiments, $N_A=N_B=N_C/2=N$.

\subsection{Robustness to Sampling}
\begin{figure*}[t]
\centering
\begin{subfigure}{.32\textwidth}
    % \centering
\includegraphics[width=0.98\linewidth]{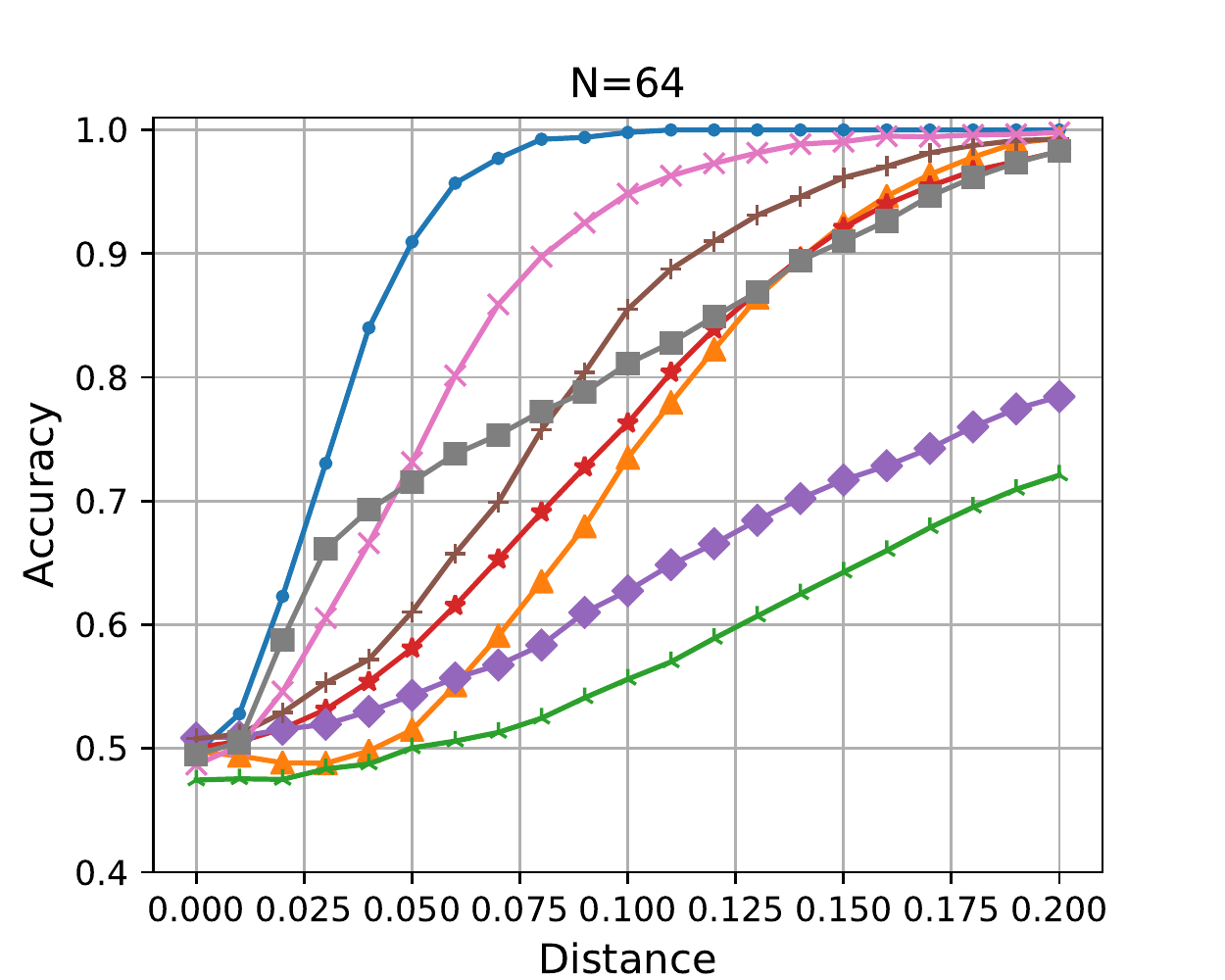}
% \caption{}
    \label{fig:fig_move_testchair64}
\end{subfigure}
\begin{subfigure}{.32\textwidth}
    % \centering
    \includegraphics[width=0.98\linewidth]{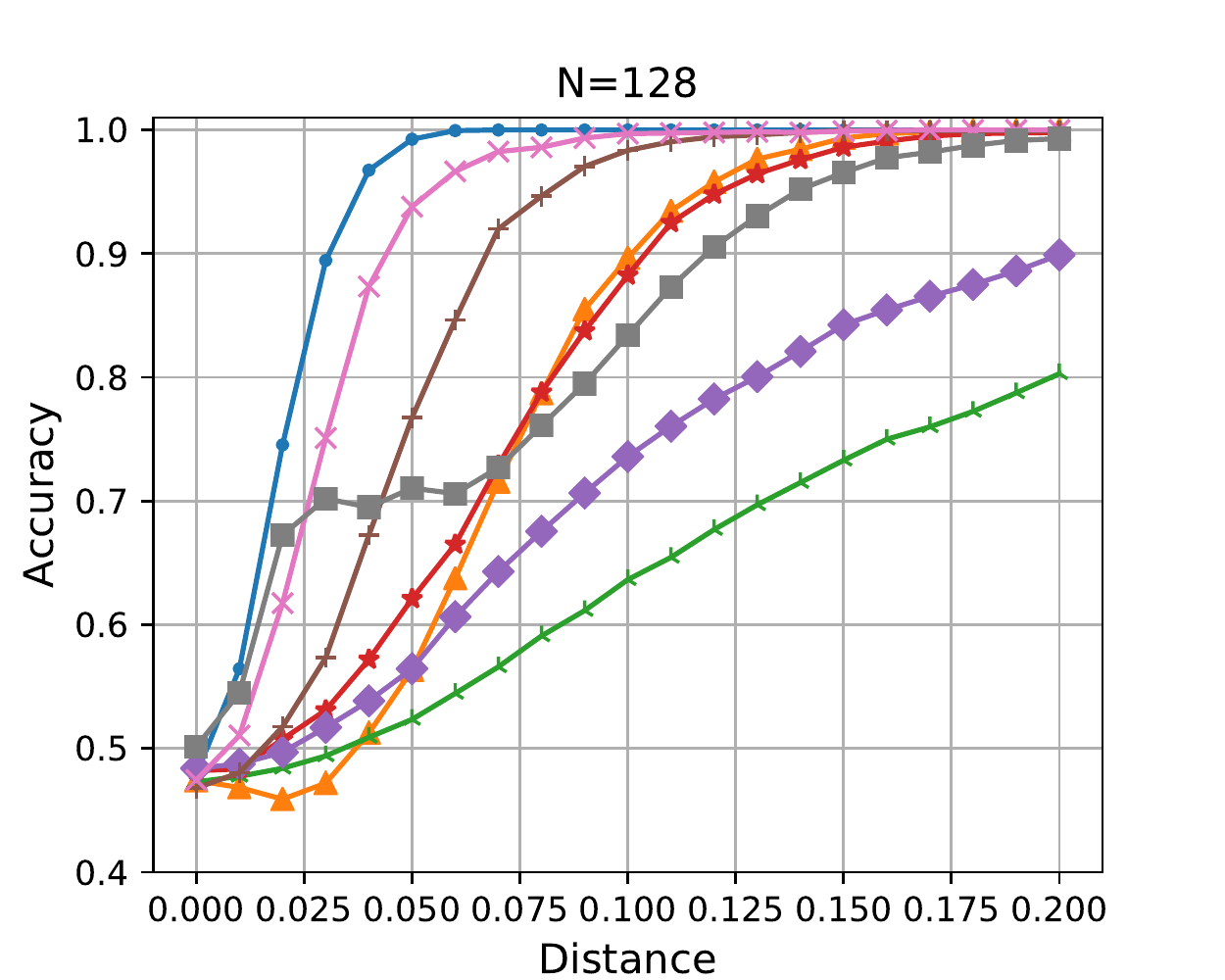}
    % \caption{}
    \label{fig:fig_move_testchair128}
\end{subfigure}
\begin{subfigure}{.32\textwidth}
    % \centering
    \includegraphics[width=0.98\linewidth]{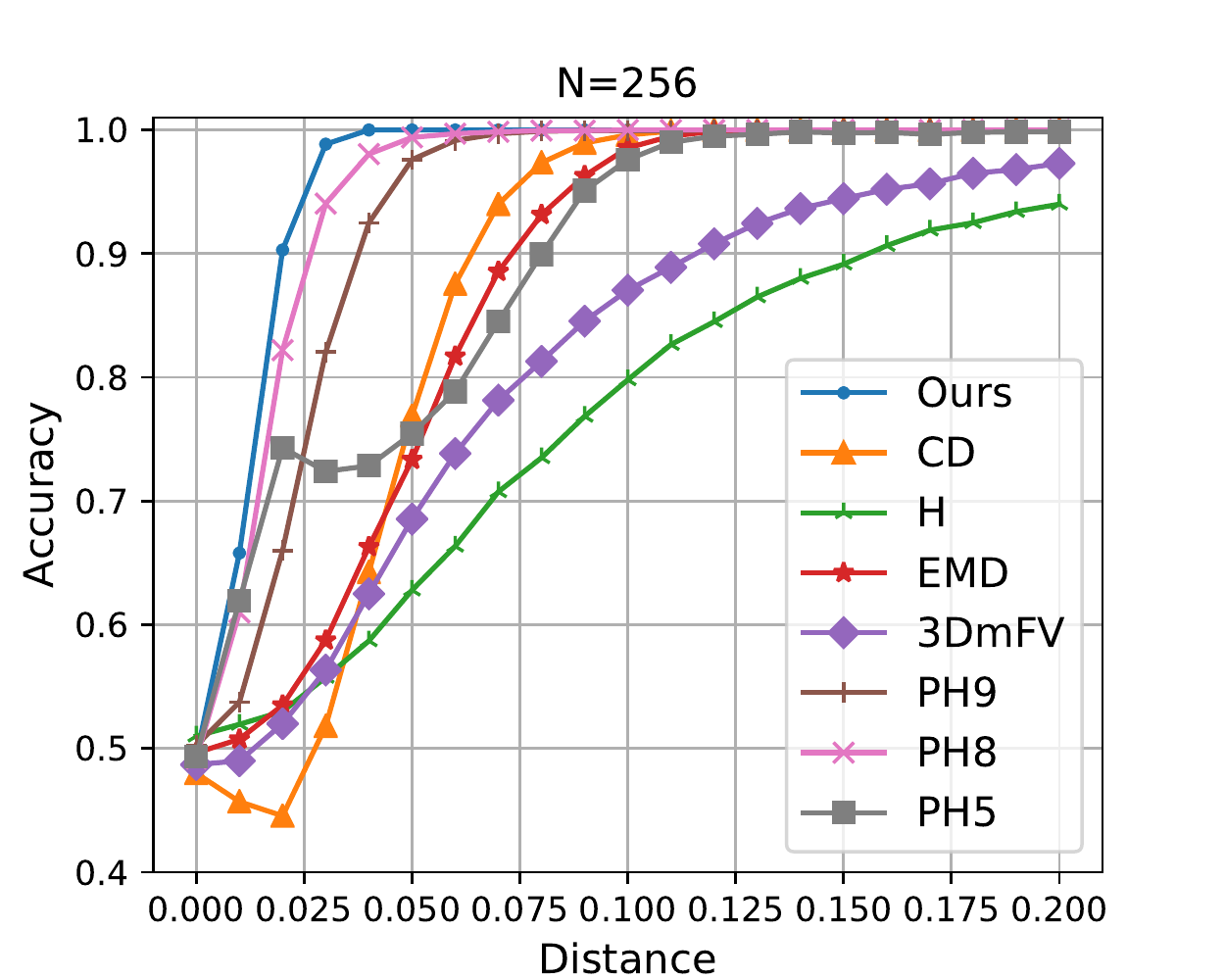}
    % \caption{}
    \label{fig:fig_move_testchair256}
\end{subfigure}
\begin{subfigure}{.32\textwidth}
    % \centering
\includegraphics[width=0.98\linewidth]{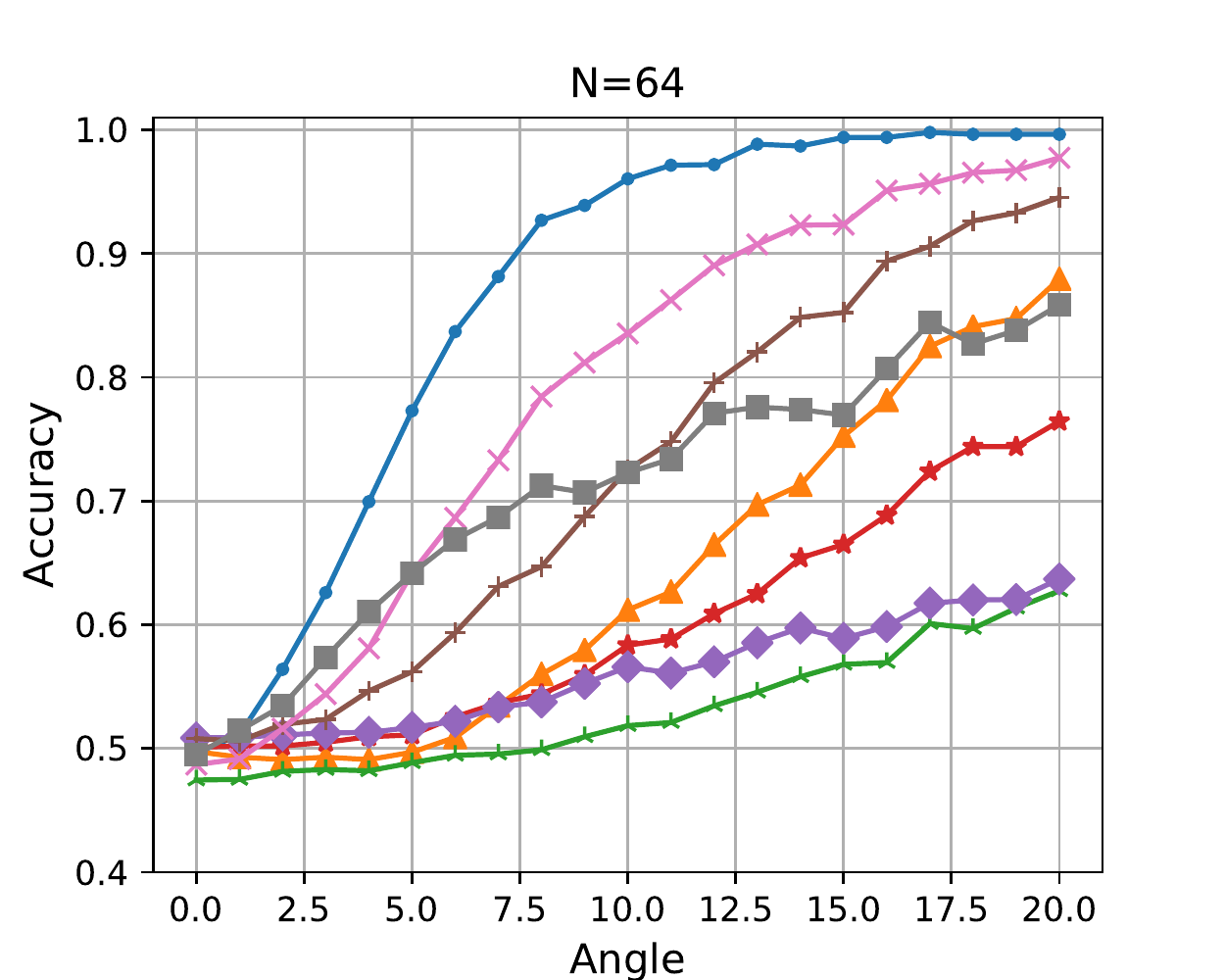}
% \caption{}
    \label{fig:fig_rotate_testchair64}
\end{subfigure}
\begin{subfigure}{.32\textwidth}
    % \centering
    \includegraphics[width=0.98\linewidth]{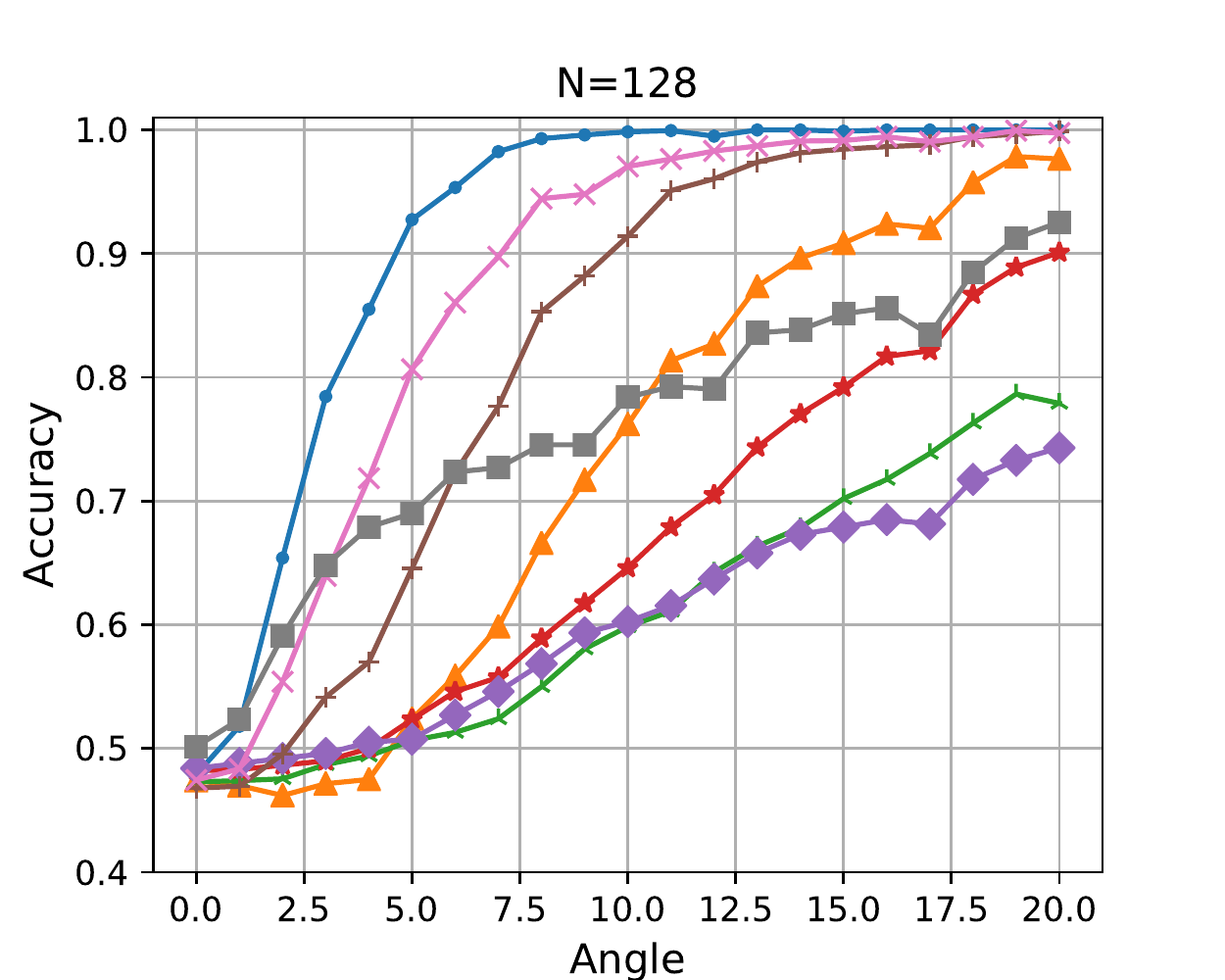}
    % \caption{}
    \label{fig:fig_rotate_testchair128}
\end{subfigure}
\begin{subfigure}{.32\textwidth}
    % \centering
    \includegraphics[width=0.98\linewidth]{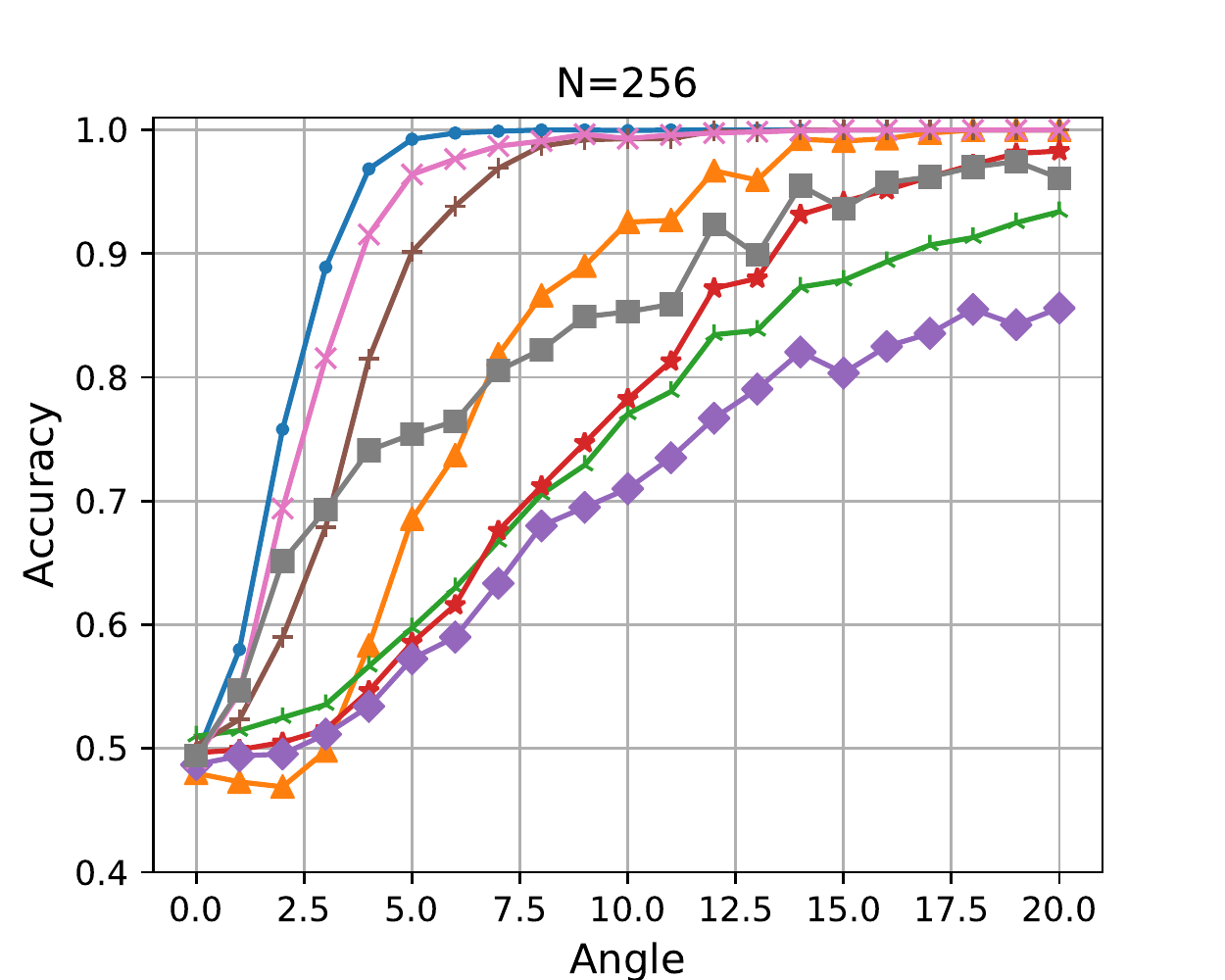}
    % \caption{}
    \label{fig:fig_rotate_testchair256}
\end{subfigure}
\caption{Detecting translation (top) and rotations (bottom) using DPDist and other point cloud distances, for different point cloud densities $N=64,128,256$.
% We can see that DPDist achieve higher accuracy than CD and EMD.
}
\label{fig:mov_test}
\end{figure*}

Discriminating between very different objects from dense samples of their surfaces is easy, and performance is highly independent of the actual sampling. Harder tasks and sparser sampling are more challenging. 
We consider here three discrimination tasks associated with similar objects, perform them using different distances between their point cloud samples, and examine  their robustness to sampling.

We conducted the experiments on the "Chair" category test-set from Modelnet40 with a total of 100 chairs.

% \subsubsection{Translation detection test:}
\noindent\textbf{Translation detection test:}
The task here is to discriminate between a stationary object and a moving object. Thus, in every test we consider three point clouds: $S_A$ - a sampling of the object in its original position, $S_B$ - a sampling of the object after it was translated (see below), and $S_C$ - another sampling of the same object without movement. We can discriminate between the stationary object and the moving object, when the following  distance inequality
${\cal D}(S_C,S_A) < {\cal D}(S_B,S_A)$ holds true. 
% If it holds
% , we can make the correct decision using the point clouds and the distance. 
We consider translation along the 26 directions  $(\theta_1,\theta_2,\theta_3), \theta_{1,2,3}\in \{0,1,-1\}$, with magnitude uniformly sampled in $[0,0.2]$. 
% on the first point cloud: $S_A+t$ ($t\in \mathbb{R}^3, |t|\in[0,0.2]$). 
The accuracy is defined separately for each translation magnitude, as a fraction of successful tests. The results are shown in Fig. \ref{fig:mov_test} (top). As expected, for large translations and high sampling density, almost all methods succeed. For small densities and translations, some distances succeed better and the proposed method is best by a large margin. Remarkably, the partial Hausdorff variants succeed better than CD and EMD. 
% \subsubsection{Rotation detection test:}
\\
\noindent\textbf{Rotation detection test:}
This time, the task is to discriminate between a stationary object and a rotated object.
% We repeat the last experiment, and we switch the translation with rotations.
The rotation angle magnitude is uniformly sampled in $[0,20]$ (deg) around the 26 directions specified in the last section. Fig. \ref{fig:mov_test} (bottom) shows
that DPDist performs significantly better than the other distances in the presence of small rotations and densities. \\
% \subsubsection{Identification test:}
\noindent\textbf{Identification test:}
\begin{figure*}[t]
% \centering
\begin{subfigure}{.31\textwidth}
    % \centering
\includegraphics[width=0.98\linewidth]{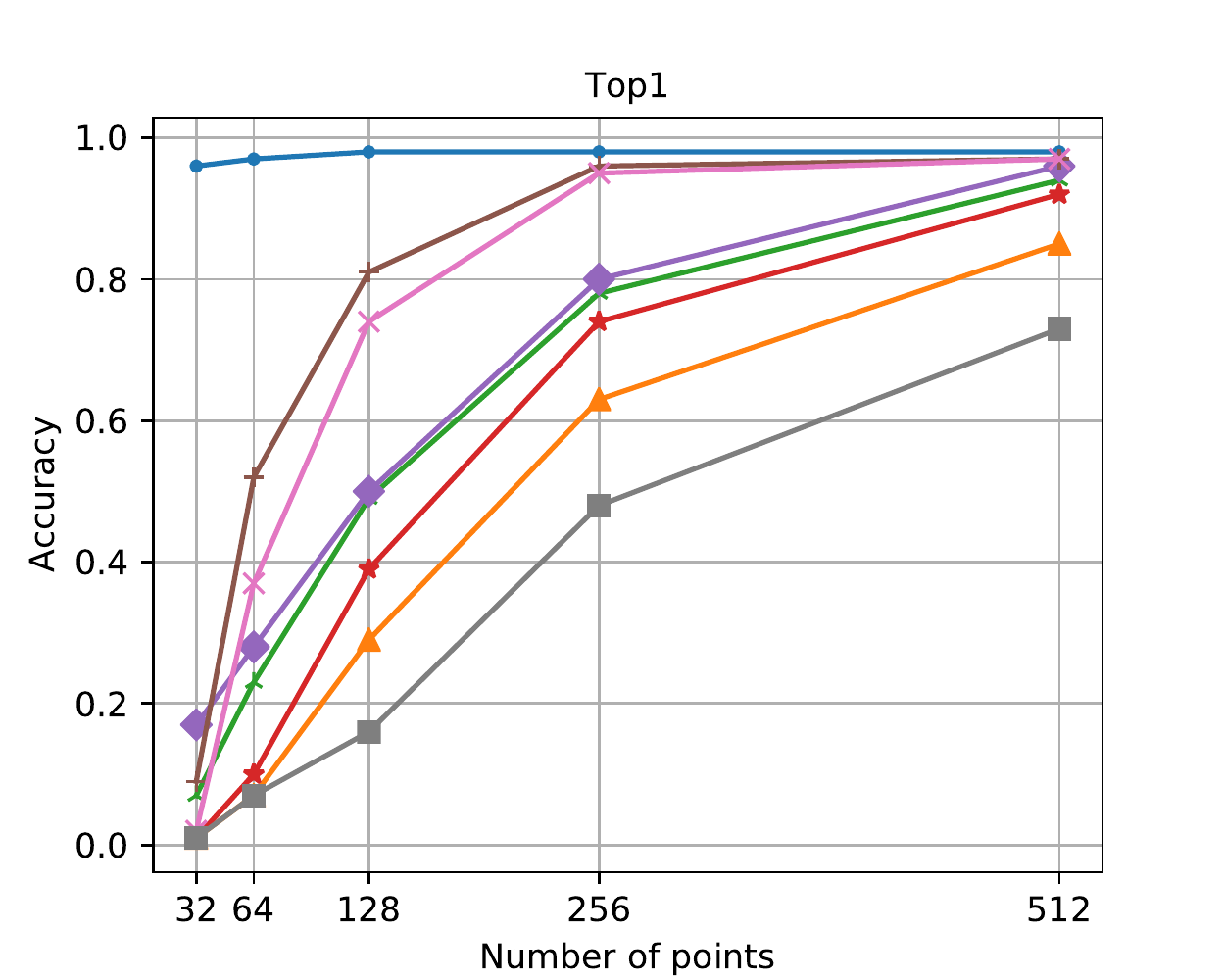}
% \caption{}
    \label{fig:points_retTop1}
\end{subfigure}
\begin{subfigure}{.31\textwidth}
    % \centering
    \includegraphics[width=0.98\linewidth]{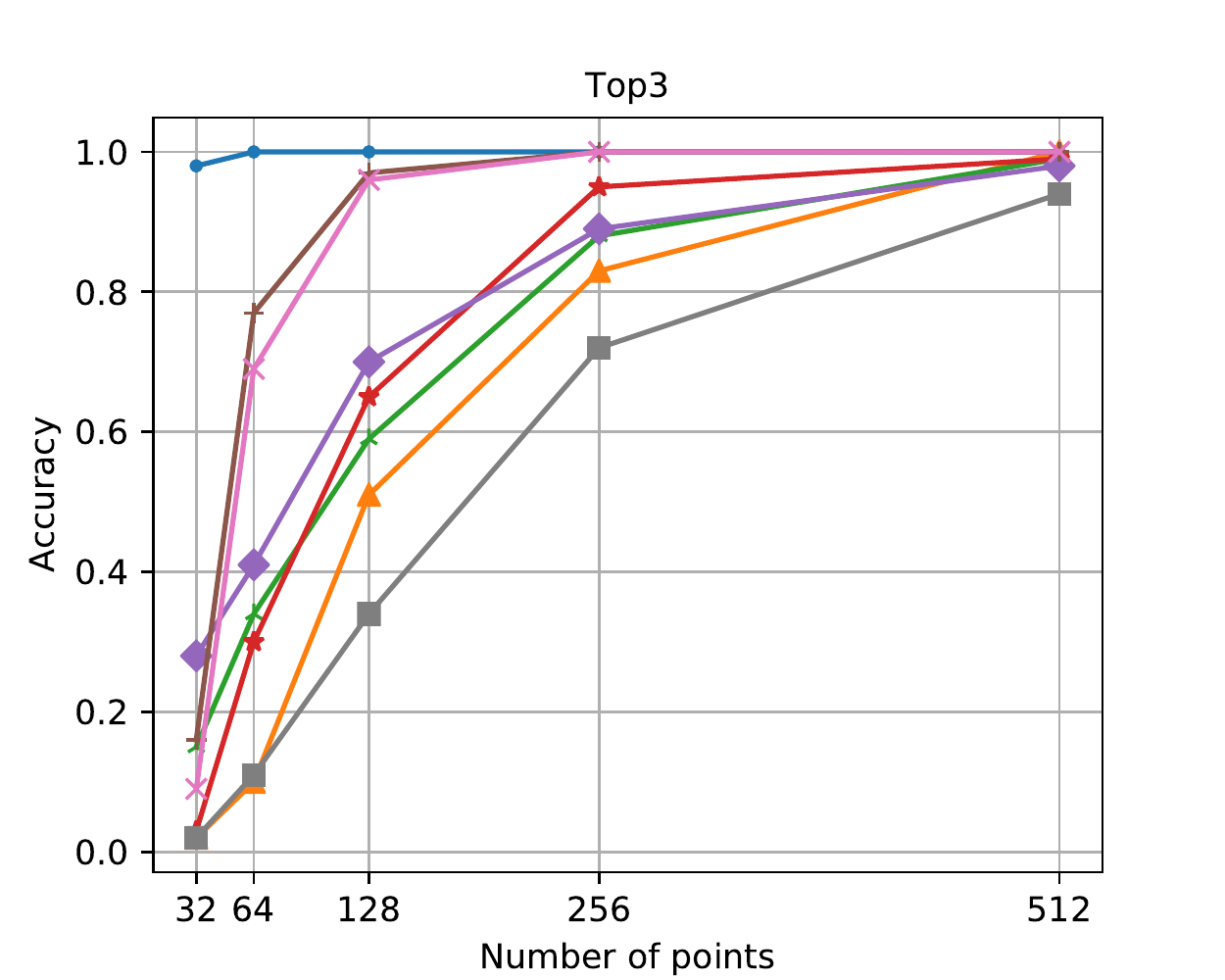}
    % \caption{}
    \label{fig:points_retTop3}
\end{subfigure}
\begin{subfigure}{.31\textwidth}
    % \centering
    \includegraphics[width=0.98\linewidth]{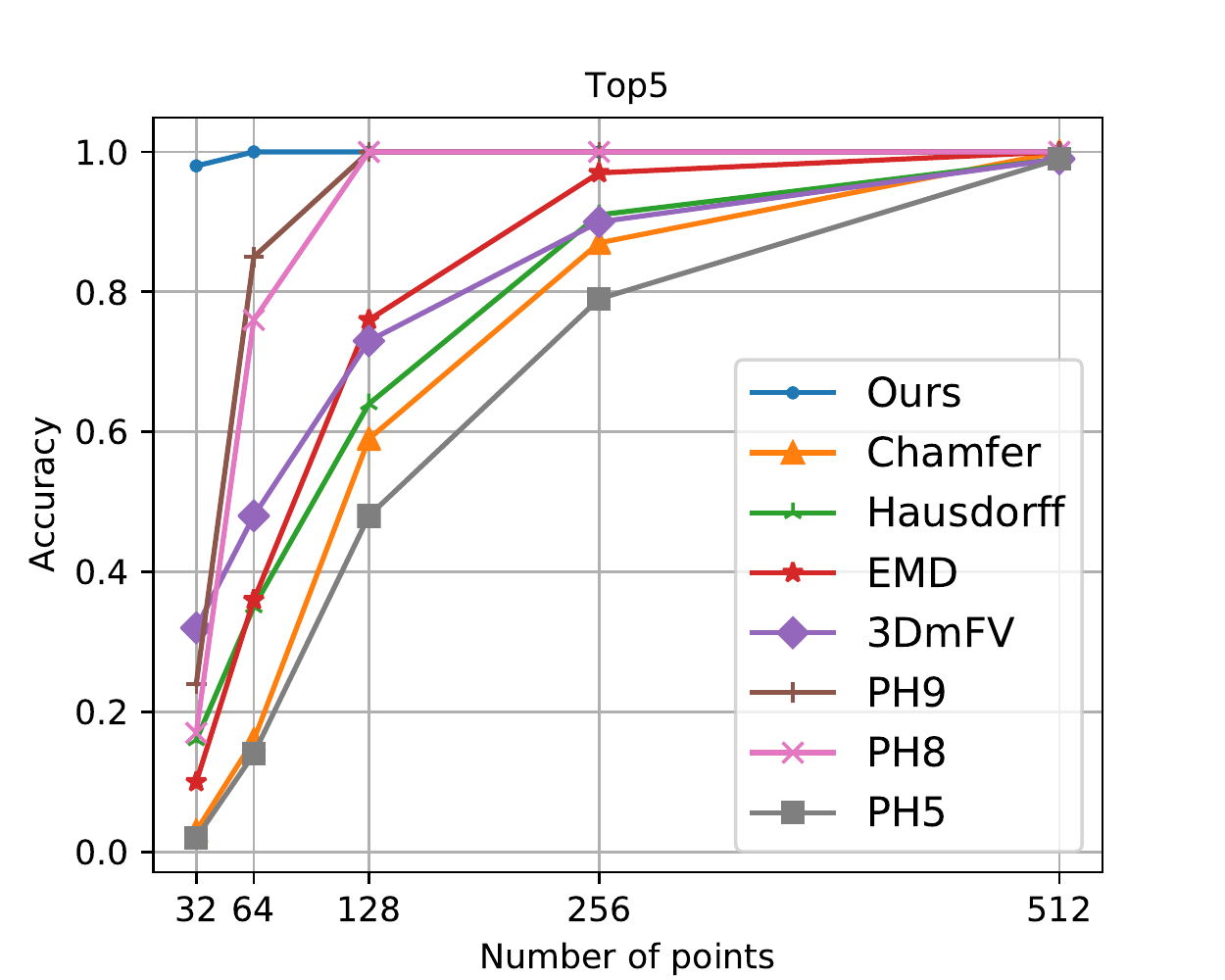}
    % \caption{}
    \label{fig:points_retTop5}
\end{subfigure}
\caption{Instance identification  test - comparing top 1, 3, and 5 accuracy in nearest neighbor identification robustness test. 
The DPDist is the clear winner.
}
\label{fig:points_retTop}
\end{figure*}
We evaluate the sampling robustness of different point cloud distances by testing their ability to discriminate between objects from the same category. Given two samples of the same CAD object $S_A$ and $S_A'$,  we test if the two samples of the same object are closer to each other than to samples of any other objects from the same category.
For each point cloud $S_A$, we sort its distances from $S_A'$ and from the point cloud of all other objects. 
The identification is considered "Top $m$" successful if $S_A'$ is in the "$m$" closest point clouds to $S_A$ (out of 100).  
The success rate is the fraction of objects for which the identification is "Top $m$" successful. Fig. \ref{fig:points_retTop} shows that the proposed DPDist method is robust and works much better than other methods. Remarkably, the partial Hausdorff distance is the next best method, and it easily overcomes the more standard CD and EMD. 

\subsection{Local Representation Learning Generalization}

% \subsection{Local Representation Learning Generalization}
\begin{figure*}
\centering
\begin{subfigure}{.32\textwidth}
    \centering
\includegraphics[width=0.98\linewidth]{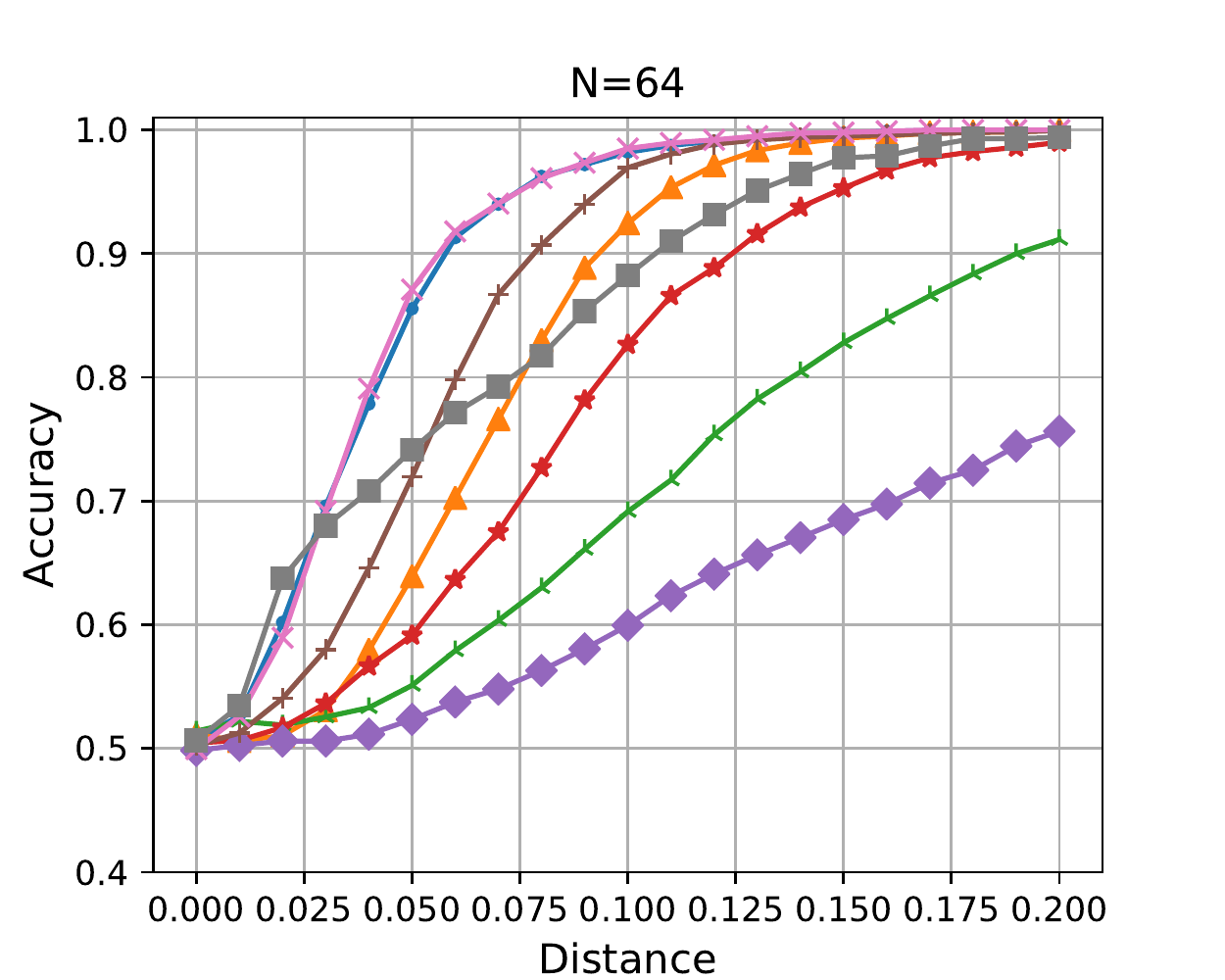}
% \caption{Airplane}
    \label{fig:}
\end{subfigure}
\begin{subfigure}{.32\textwidth}
    \centering
\includegraphics[width=0.98\linewidth]{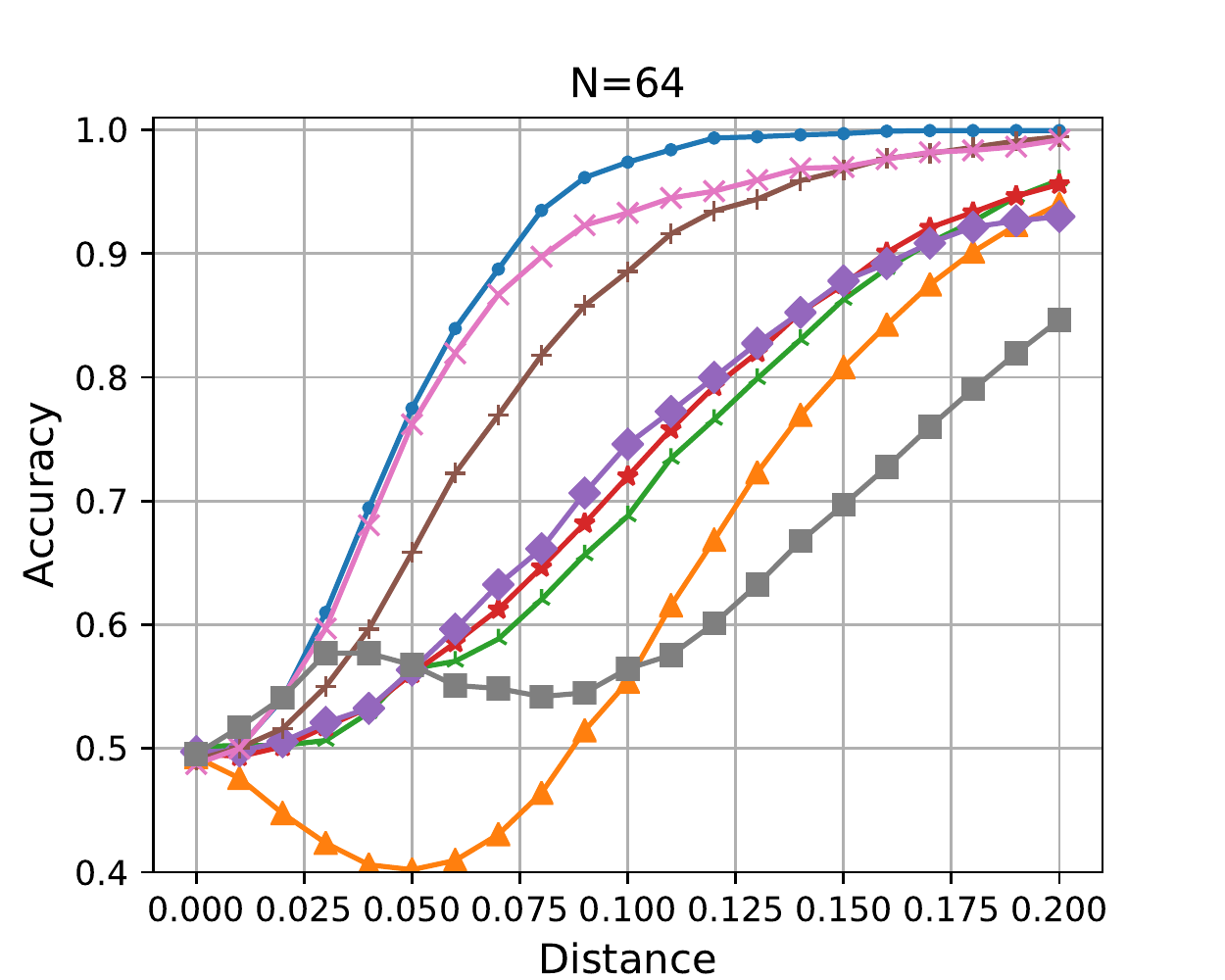}
% \caption{Car}
    \label{fig:}
\end{subfigure}
\begin{subfigure}{.32\textwidth}
    \centering
\includegraphics[width=0.98\linewidth]{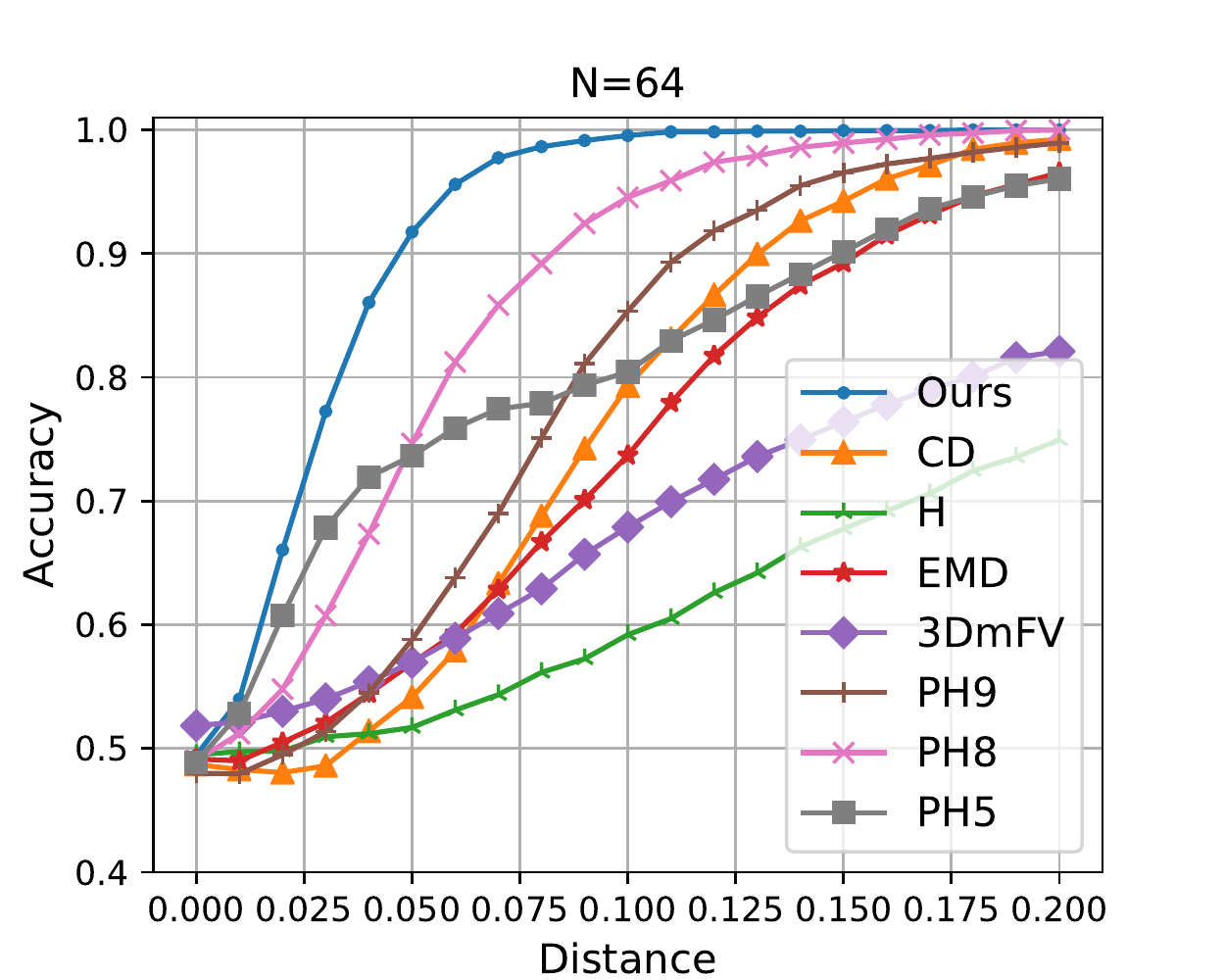}
% \caption{Table}
    \label{fig:}
\end{subfigure}
\begin{subfigure}{.32\textwidth}
    \centering
\includegraphics[width=0.98\linewidth]{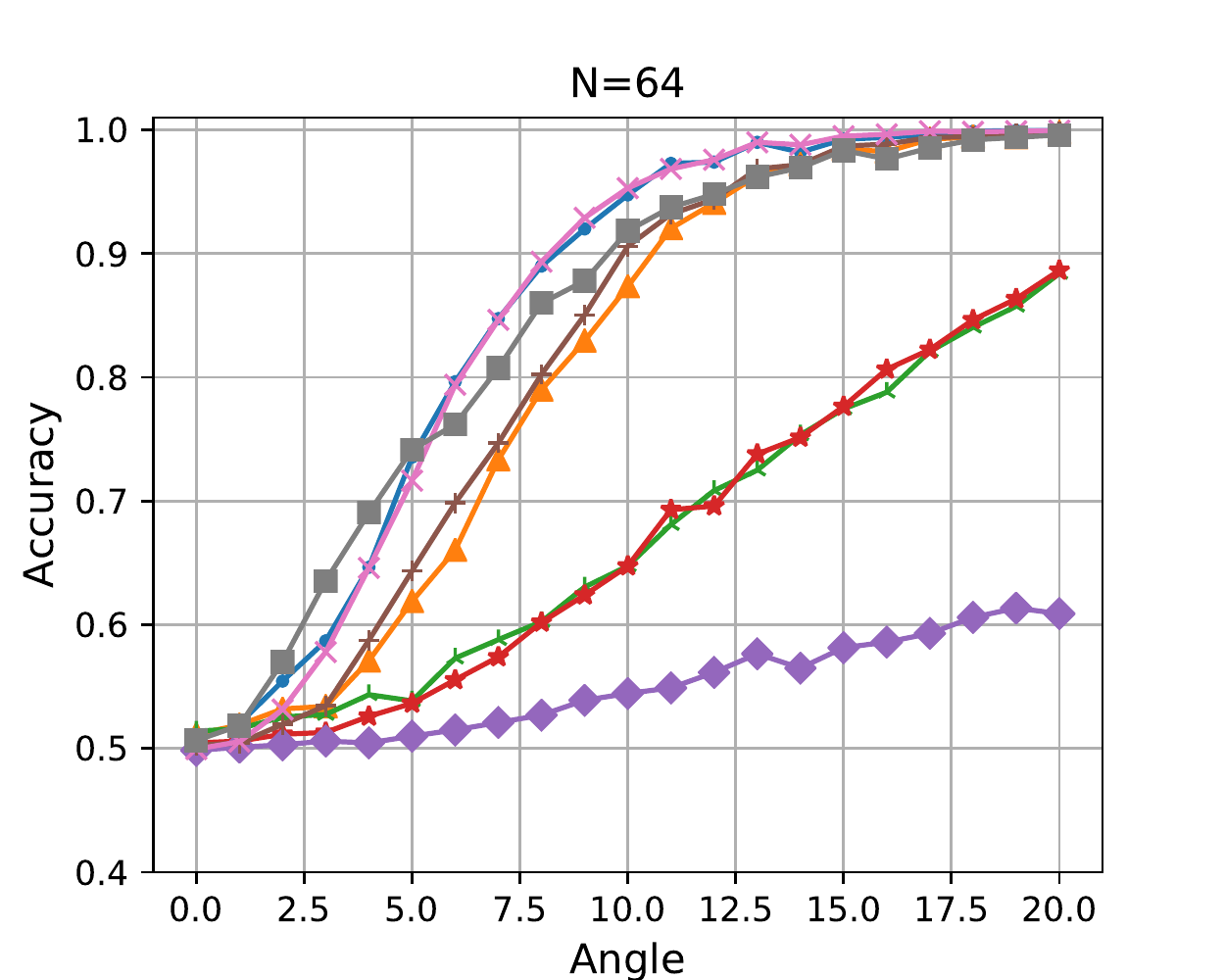}
\caption{Airplane}
    \label{fig:}
\end{subfigure}
\begin{subfigure}{.32\textwidth}
    \centering
\includegraphics[width=0.98\linewidth]{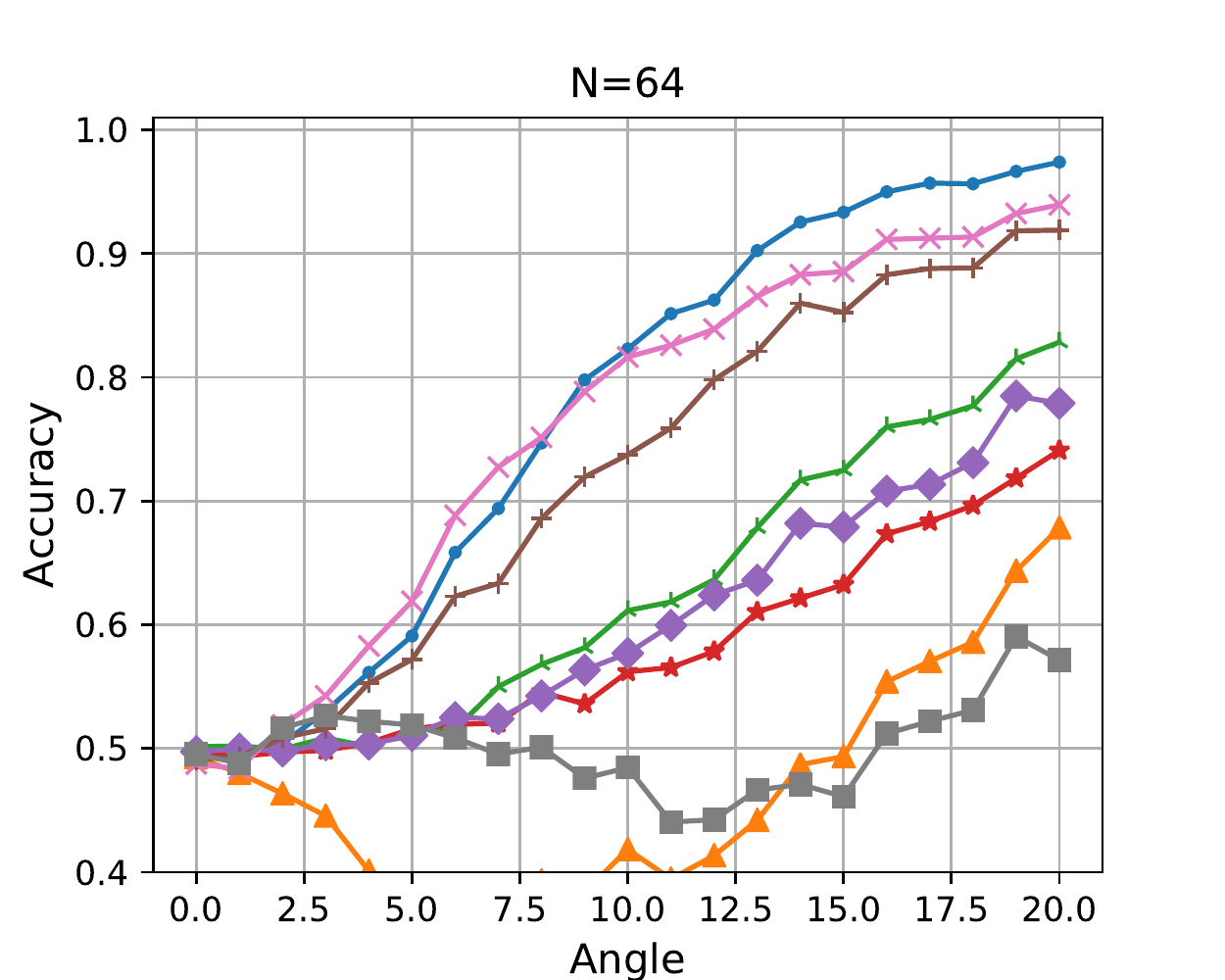}
\caption{Car}
    \label{fig:}
\end{subfigure}
\begin{subfigure}{.32\textwidth}
    \centering
\includegraphics[width=0.98\linewidth]{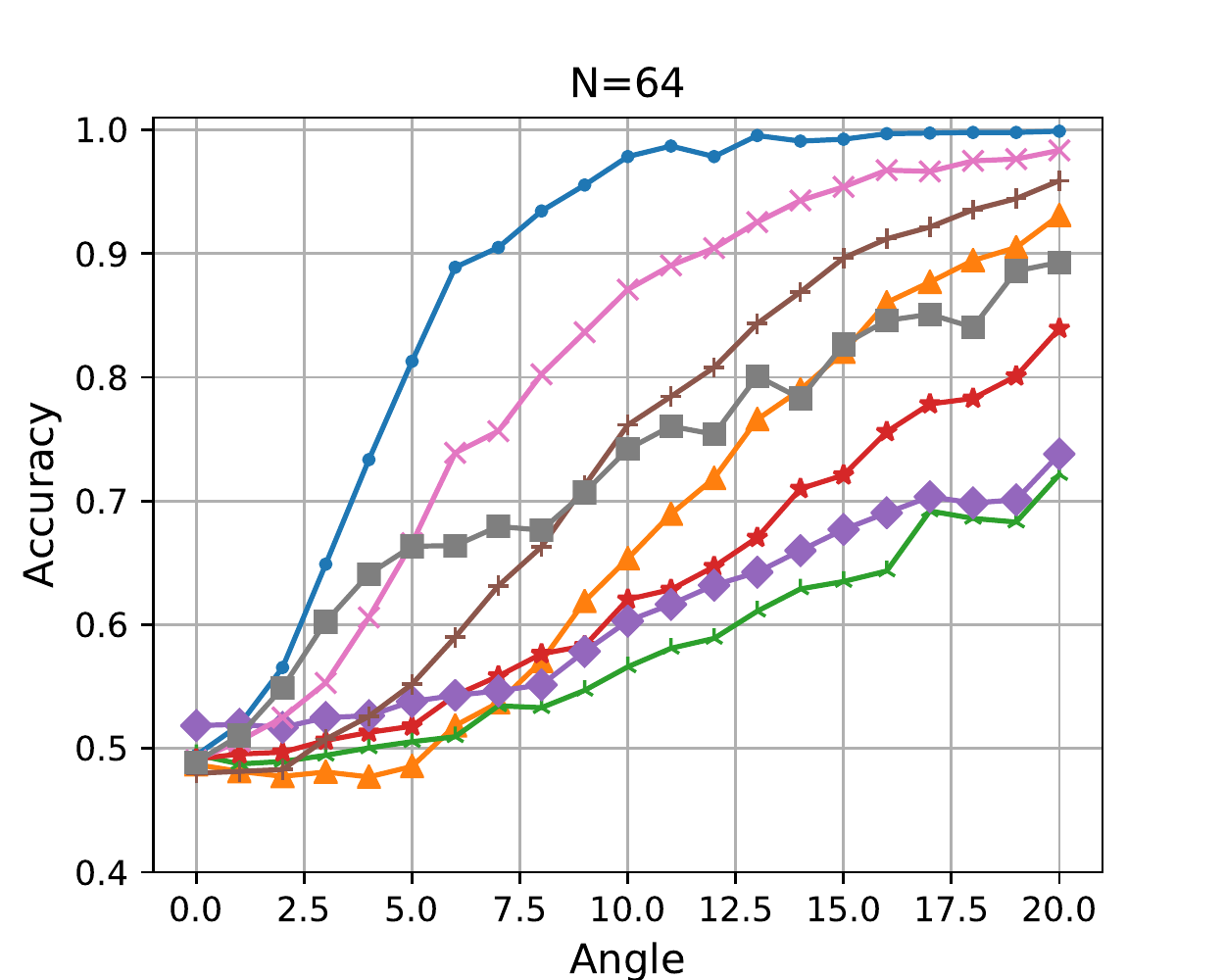}
\caption{Table}
    \label{fig:}
\end{subfigure}
\caption{Translation (top) and rotation (bottom) detection experiments evaluated on (a) "Airplane" (b) "Car", and (c) "Table" as before. Remarkably, DPDist is still the best method, although the SPD was  trained over another category ("Chair") and not over these categories. The generalization ability is a result of using local surface modelling.   }
\label{fig:results:local_presentation_rubastness}
\end{figure*}
A major advantage of the proposed method is its ability to learn local representations, which makes the implicit surface estimation effective and computationally efficient. 

Objects are different from each other but often share local parts. That is, many objects contain planes, corners, curved surfaces, etc. We therefore hypothesize that training the SPD network over local representations of just one reasonably rich object category (e.g., "Chair") makes the learning universal, because the category includes enough variations of small patches to generalize for unseen categories.
Therefore, although we train our SPD only on the "Chair" category, we expect it to generalize to other categories. 

We test our hypothesis by performing the translation and rotation tests over "Airplane", "Car", and "Table" categories on a network that was trained only on "Chair".
Fig. \ref{fig:results:local_presentation_rubastness} shows that the DPDist is still the most discriminative method and more robust to sampling than the other tested methods.

\subsection{Applying Point Cloud Distances to Registration Learning}
\begin{figure*}[t]
    \centering
    \includegraphics[width=0.95\linewidth]{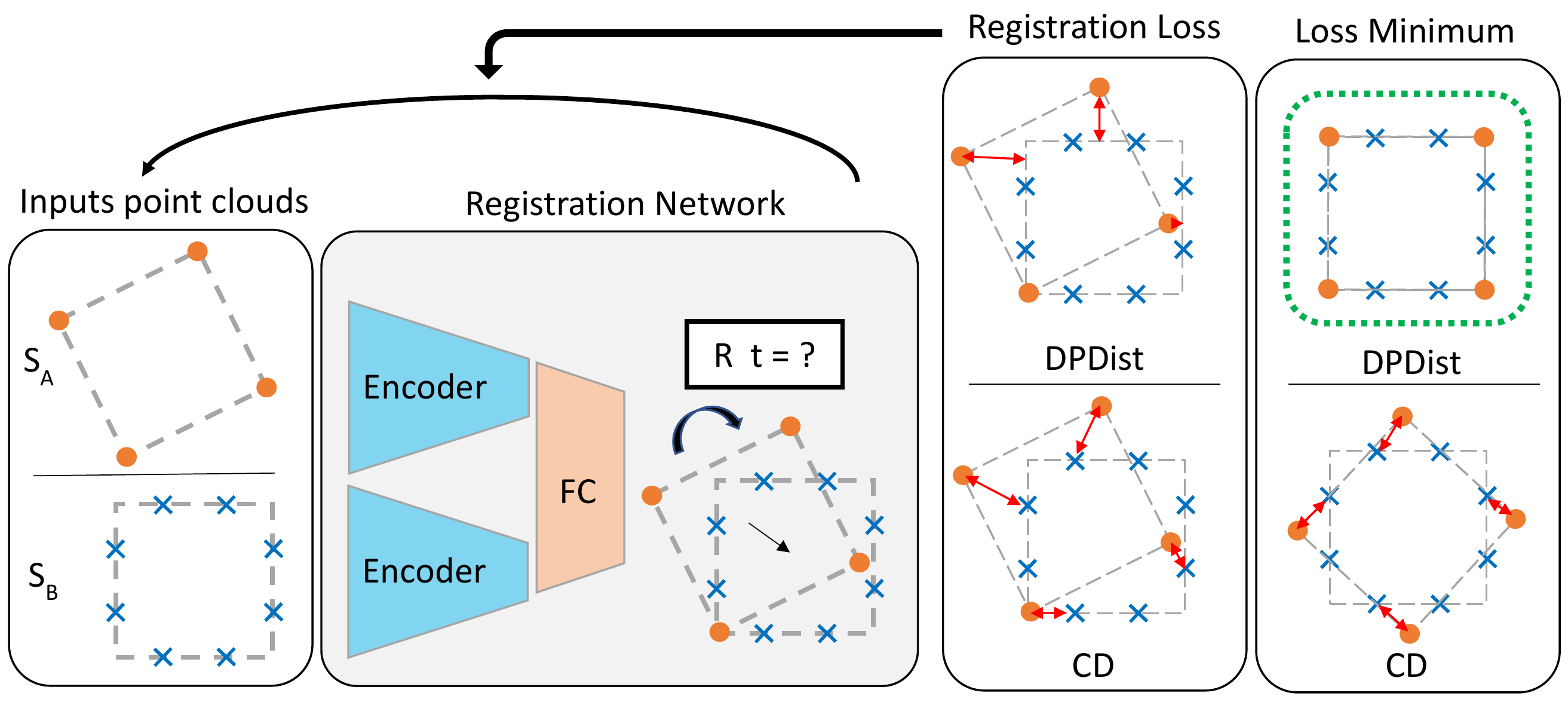}
    \caption{Learning registration using DPDist vs. CD distances as loss function. Two point cloud ($S_A$, $S_B$) are input to a network that regresses the transformation between them. The chosen loss highly effects the output transformation as the DPDist loss minimum is consistent with zero distance, while Chamfer loss results in nonzero distance for different sampling of the same object.}
    \label{fig:DPDist_reg_illustration}
\end{figure*}

In this section, we use an existing registration network, and show its performance improvement 
% when using DPDist as a distance between point clouds. 
when using DPDist as its loss function.
We choose the recent
PCRNet \cite{sarode2019pcrnet} (see also Section \ref{rw:reg}) since it compares point clouds in its training loss function. 
Fig. \ref{fig:DPDist_reg_illustration} illustrates the challenges of using a correspondence-based distance vs. the proposed distance.
We compare DPDist to CD and EMD distances as the loss function for training the iterative PCRNet.

\begin{figure*}[h]
\centering
\begin{subfigure}{.22\textwidth}
    \centering
\includegraphics[width=0.98\linewidth]{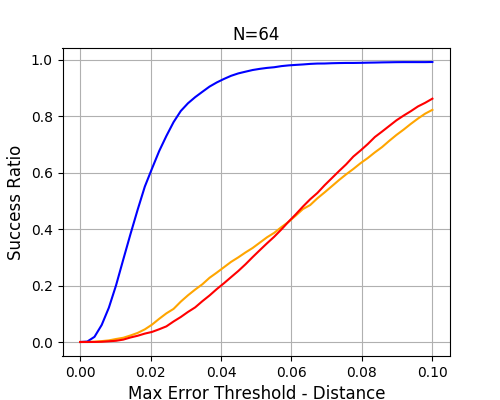}
% \caption{}
    \label{fig:fig_rotate_testchair64}
\end{subfigure}
\begin{subfigure}{.22\textwidth}
    % \centering
    \includegraphics[width=0.98\linewidth]{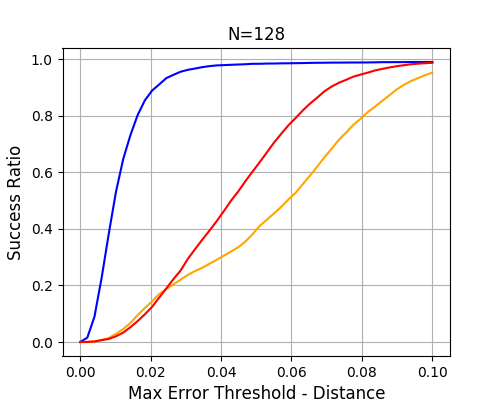}
    % \caption{}
    \label{fig:fig_rotate_testchair128}
\end{subfigure}
\begin{subfigure}{.22\textwidth}
    % \centering
    \includegraphics[width=0.98\linewidth]{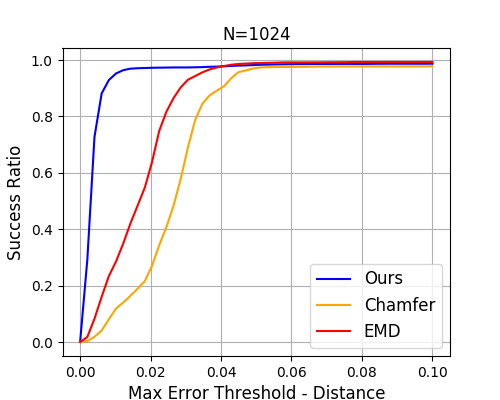}
    % \caption{}
    \label{fig:fig_rotate_testchair256}
\end{subfigure}
\begin{subfigure}{.22\textwidth}
    % \centering
    \includegraphics[width=0.98\linewidth]{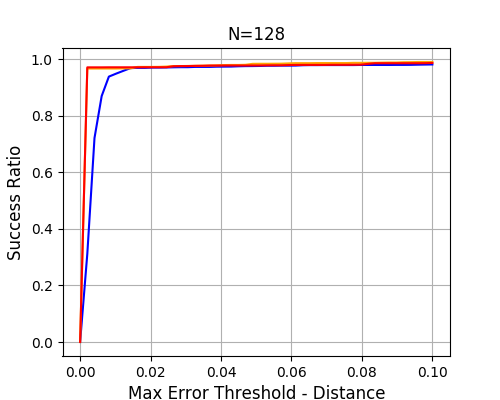}
    % \caption{}
    \label{fig:fig_rotate_testchair256}
\end{subfigure}
\begin{subfigure}{.22\textwidth}
    \centering
\includegraphics[width=0.98\linewidth]{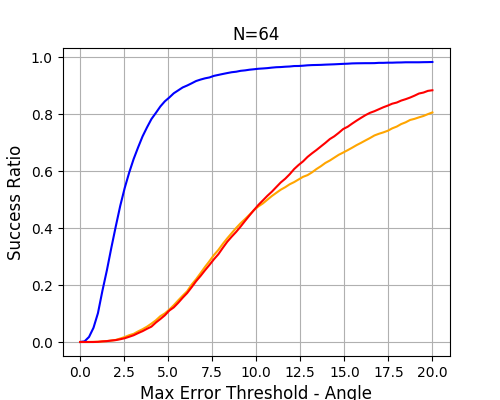}
\caption{}
    \label{fig:fig_rotate_testchair64}
\end{subfigure}
\begin{subfigure}{.22\textwidth}
    % \centering
    \includegraphics[width=0.98\linewidth]{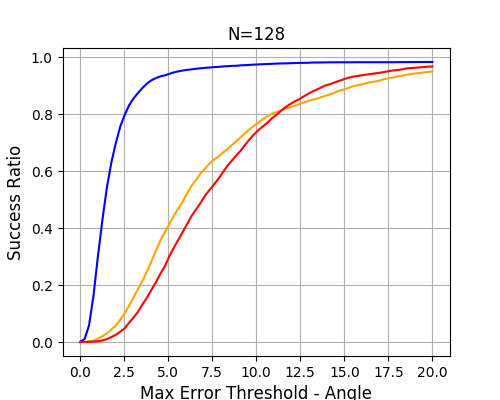}
    \caption{}
    \label{fig:fig_rotate_testchair128}
\end{subfigure}
\begin{subfigure}{.22\textwidth}
    % \centering
    \includegraphics[width=0.98\linewidth]{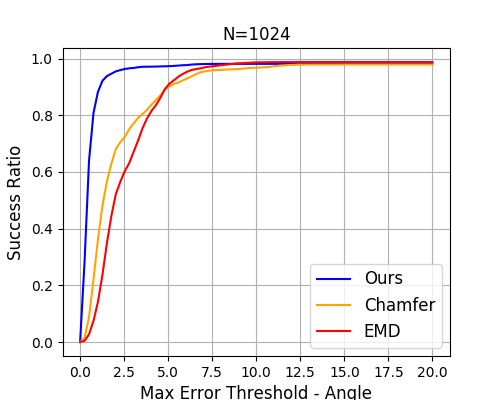}
    \caption{}
    \label{fig:fig_rotate_testchair256}
\end{subfigure}
\begin{subfigure}{.22\textwidth}
    % \centering
    \includegraphics[width=0.98\linewidth]{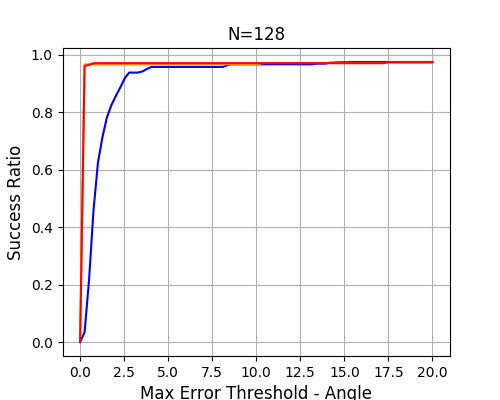}
    \caption{}
    \label{fig:fig_rotate_testchair256}
\end{subfigure}
\caption{Results for training PCRNet using CD, EMD, and the proposed DPDist over a single category ("Chair"). 
We show the success ratio vs. max error threshold for translation (top) and rotation (bottom) of the following input point cloud sizes: $N=64, 128, 1024$ (a,b,c). 
DPDist consistently achieves high performance for various densities.
(d) shows the case of two identical samples $S'_A=S_B$, where CD and EMD gets nearly perfect results thanks to the high correspondence between the clouds' samples.
}
\label{fig:results:registration_dif_num_point}
\end{figure*}
% \subsubsection{Metric evaluation:} 
% In this section we evaluate our distance metric as a loss function for the registration framework of the iterative PCRNet.
We used the PCRNet as described in \cite{sarode2019pcrnet}, but found that limiting the output rotation angle to 45 degrees improves the stability when training on small point clouds. In the context of iterative registration, this modification does not limit the registration range. 
Using the "Chair" category and following  \cite{sarode2019pcrnet}, we randomly generate 5070 different transformations for training and other 5070 transformations for testing. The transformations include a rotation between $[-45,45]$ degrees about one random direction and a translation between $[-0.1,0.1]$ in another random direction. The evaluation metric at test time is the "success ratio"  \cite{sarode2019pcrnet}: the percentage of point clouds with a transformation error under a given 'max error' threshold. 

To demonstrate the DPDist robustness, we conduct the following experiment: Given a CAD model input, we generate two point clouds  by sampling $2N$ points using FPS, and randomly divide it into two disjoint clouds $S'_A$ and $S_B$ of size $N$. We transform $S'_A$ to $S_A$ and store the transformation matrix as ground truth. We then train PCRNet using different distance losses. Fig. \ref{fig:results:registration_dif_num_point} shows the results for training PCRNet using CD, EMD, and the proposed DPDist for different sample sizes. The significant advantage of DPDist is clear. 

The relative inaccuracy of the CD and EMD distances is due to the different samples. To demonstrate that this is indeed the reason, we added another success ratio curve with $128$ points, but this time, $S'_A=S_B$. That is, the two point clouds (before transformation) are identical. We can see in Fig. \ref{fig:results:registration_dif_num_point}d that this time, the performance with CD and EMD is nearly perfect. 

\section{Conclusions}

Comparing between point clouds is a fundamental data analysis task. For point clouds, obtained by sampling some underlying continuous object surfaces, the actual hidden goal is to compare between these objects. Most methods, however, rely on  distances between the raw points and are therefore sensitive to the uncertainties involved in the sampling process. 

The method proposed here, DPDist, estimates the distances of points from one cloud to the underlying continuous surface corresponding to the other point cloud and the other way around. 
DPDist is fast and effective. It is more accurate than the commonly used Chamfer distance and Earth mover's distance methods. Its advantage is significant especially for difficult tasks, such as discriminating between similar objects, and for sparse point clouds. For example, using a DPDist dependent loss function, we were able to train a registration network so that it provides good results with point clouds of size 64. 

Unlike other methods, we provide a fast process for generating the surface representation. This representation is local but is created efficiently as part of a global representation. Being local enables us to represent fine surface details with a moderately deep network, and makes the learned network universal: learning the local descriptions associated with one category is general enough to represent local descriptions of objects from other categories. 
\\
\\
% \section*{Acknowledgments}
\noindent\textbf{Acknowledgments:}
This research was supported by the Israel Science Foundation, and by the Israeli Ministry of Science and Technology.
\clearpage
\bibliographystyle{splncs04}
\bibliography{egbib}
\clearpage

\section*{Supplementary Material}

\subsection{Training details} %Additioanl?

The SPD network is trained with a batch size of 16 for 1000 epochs, using the Adam optimizer, a learning rate of $0.001$ with an exponential decay rate of $0.5$ every $3\times10^5$ steps, and batch normalization.
Training the SPD network takes approximately 3 hours ($N=256$), and using DPDist to measure the distance between point clouds takes 6ms. 
All timing approximations were tested on NVIDIA GeForce RTX 2080 Ti GPU and an Intel Core i9 CPU at 3.60GHz.

\subsection{Ablation study}

We explore the influence of the 3DmFV parameters (number of local and global Gaussians) on our method's performance. 
Given an input point cloud of 512 points, we reconstruct a mesh from the learned implicit function representation using Marching cubes \cite{lorensen1987marching}. We then sample 10k points from the reconstructed mesh and compare them to 10k points sampled on the original CAD model using Chamfer L1 and normal consistency as specified in  \cite{mescheder2019occupancy}. For each point from one set, we find it's nearest neighbor in the second set. Then we compute the euclidean distances and the normal consistency between them and average over all points in the set. We then alternate sets and average between the two results. 
We evaluate our results on ModelNet40 datase's "chair" category.

In the first experiment we explore the influence of the global Gaussian grid size. We use a local grid size of $3^3$ and a global size of $4^3, 8^3$, and $16^3$. The results in Table \ref{table_ng} are consistent with the thorough hyper parameter study conducted in \cite{ben20183dmfv} and show that $8^3$ Gaussian grid is adequate, balancing the computation-accuracy trade-off.

In the second experiment we explore the influence of the local Gaussian grid size. We use a global grid size of $8^3$ and a local size of $1^3, 3^3$, and $5^3$. The results in Table \ref{table_ps} show that the results between $3^3$ and $5^3$ are comparable with a slight advantage to $3^3$. However, this is most likely attributed to the small size of the dataset and we chose to use $5^3$ in our experiments, maintaining a higher network capacity. 

\begin{table}[h]
\centering
\begin{tabular}{ |p{4.0cm}|p{1.8cm}|p{1.8cm}|p{1.8cm}|}
 \hline
 \textbf{Number of Gaussian} & $4\times{}4\times{}4$  & $8\times{}8\times{}8$ & $16\times{}16\times{}16$ \\
 \hline
\textbf{Chamfer L1 $\downarrow$} & 0.130189 & 0.071805 & 0.058943\\
\textbf{Normal Consistency $\uparrow$} & 0.718058 & 0.794560 & 0.809462\\
 \hline
\end{tabular}
\caption{Comparing global Gaussian grid sizes using Chamfer L1 and Normal consistency evaluation metrics. This experiment was done with a local patch size of $3^3$.
}
\label{table_ng}
\end{table}

\begin{table}[h]
\centering
\begin{tabular}{ |p{4.0cm}|p{1.8cm}|p{1.8cm}|p{1.8cm}|}
 \hline
\textbf{Local Patch Size} & $1\times{}1\times{}1$  & $3\times{}3\times{}3$ & $5\times{}5\times{}5$ \\
 \hline
\textbf{Chamfer L1 $\downarrow$} & 0.103590 & 0.071805 & 0.072714\\
\textbf{Normal Consistency $\uparrow$} & 0.693455 & 0.794560 & 0.739863\\
 \hline
\end{tabular}
\caption{Comparing local patch Gaussian grid sizes using Chamfer L1 and Normal consistency evaluation metrics.  This experiment was done with a global Gaussian grid size of $8^3$.}
\label{table_ps}
\end{table}

\subsection{Training Auto-Encoders}
In this experiment we compare DPDist to CD as loss function for training a simple auto-encoder.
We use a PointNet encoder \cite{qi2017pointnet} and three fully connected layers with sizes of $1024,1024,N*3$ as the decoder ($N$ is the number of output points). 
The loss for the auto-encoder is defined as the similarity between the output point cloud to the input point cloud. Previous works  \cite{fan2017point,achlioptas2017learning,yang2018foldingnet,groueix2018atlasnet,Li_2018_CVPR,Zhao_2019_CVPR} use the CD or EMD loss between the point clouds. 

\begin{figure*}[h]
    \centering
    \includegraphics[width=0.9\textwidth]{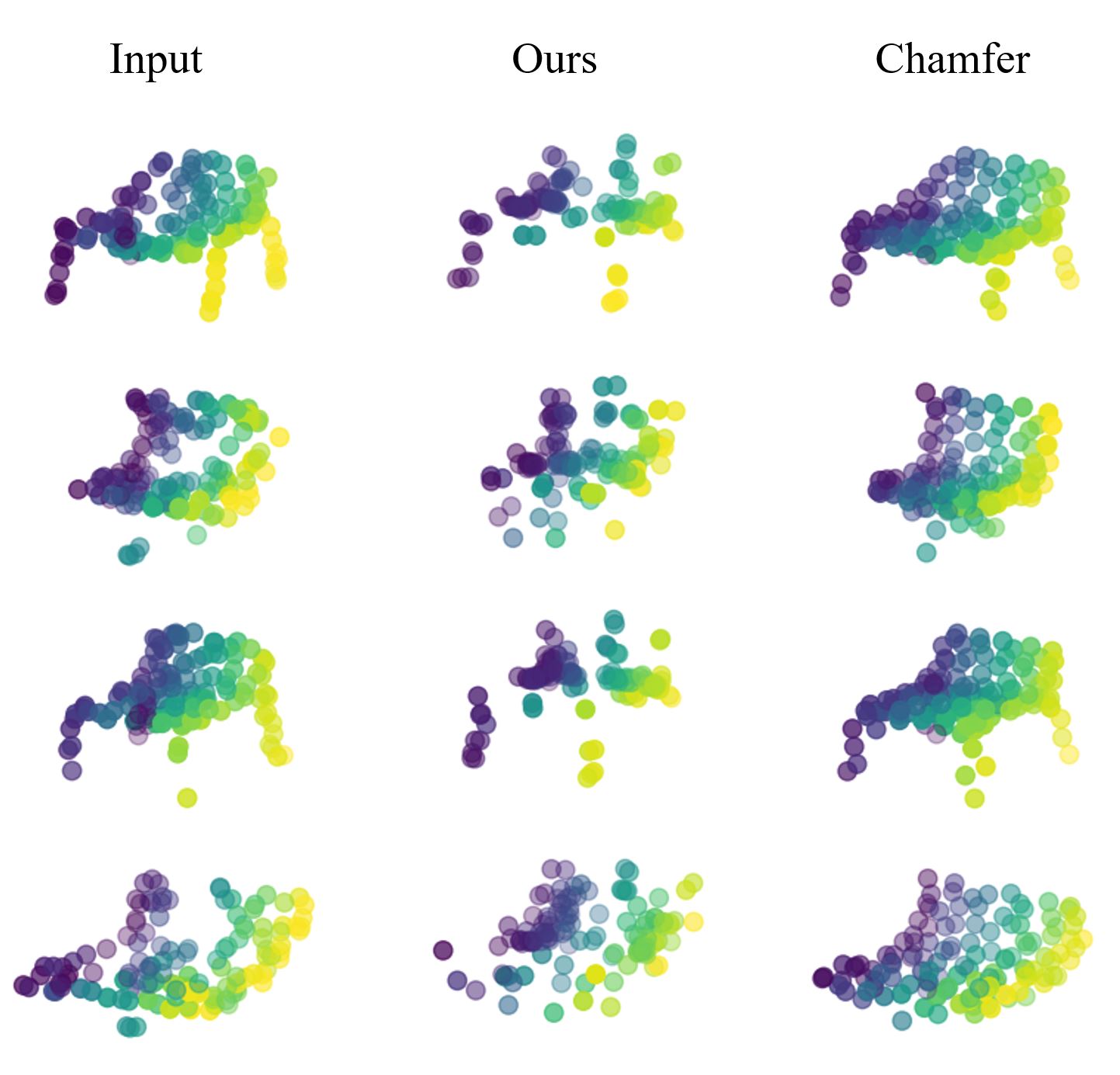}
    \caption{Point cloud auto-encoding results. We can see the auto-encoder output results when training with DPDist (middle), and Chamfer (right) distance loss for $N=128$. While Chamfer distance provides better coverage, our method gives a sparser output due to its sampling invariance property.
    }
    \label{fig:aue}
\end{figure*}

Fig. \ref{fig:aue} shows that when using DPDist, the generated point clouds suffer from high non-uniformity. Essentially, multiple points are able to coincide and satisfy the objective function.  This flaw is a direct consequence of the main strength of the proposed method: the sampling invariance property.
This property makes it robust to changes in sampling and replaces the comparison between samples to comparison between underlying surfaces. Our method is robust for both non-uniform and sparse sampling, and this is the reason it is effective in the registration task. In essence, there is a trade-off between sampling invariance and generation coverage. The focus of this paper is comparing between point clouds, therefore exploring modifications required for point generation is left for future work.

\subsection{Real-world data}
In this experiment, we use the Sydney Urban dataset, which contains LiDAR scans of outdoor objects. Because this dataset does not provide a ground truth surface, we use an equivalent class in the ModelNet dataset for training. We conduct the Translation detection test (Sec 4.2) on the car class and compare DPDist performance to the other measures. Remarkably, although the training was done on synthetic data,  our method outperforms CD, EMD, and Hausdorff and is comparable to partial Hausdorff.  Table \ref{table_real_world} reports the transformation distance where the method reached a minimum (i.e. lower is better). These results align with our CAD experiments.

\begin{table}[h]
\centering
\begin{tabular}{ |p{1.4cm}|p{1.4cm}|p{1.4cm}|p{1.4cm}| p{1.4cm}| p{1.4cm} |p{1.4cm} |p{1.4cm} |}
 \hline
 \textbf{Method} & Ours & CD & EMD & Hausdorff & PH9& PH8& PH5 \\
 \hline
 \textbf{Mean} &0.01385 &	0.02879 &	0.02381 &	0.03201 &	0.01091&0.00863&0.02807\\
 \textbf{Std.} &0.01061&  0.01119&	0.01335&		0.01391&	0.00740&0.00752&0.01491\\
 \hline
\end{tabular}
\caption{Detecting translation - given a set of translations, we report the transformation distance where the method reached a minimum (i.e., lower is better). Our method outperforms the commonly used CD and EMD, and is comparable with the partial Hausdorff variants.}.
\label{table_real_world}
\end{table}

Note that we train our method on synthetic data without data corruptions such as noise and occlusions. Further improvement may be achieved by adding more realistic scenarios into the training data or by training the proposed method directly on data collected by real-world sensors.

\end{document}